%% file: main.tex
\input{preamble}
\input{title}
\begin{document}
\input{sections/abstract}
\maketitle
\begingroup
\setlength{\parindent}{0pt}\itshape
\textbf{Corresponding Authors:}\\
Rui Yang, Department of Population Health Sciences, Weill Cornell Medicine, New York, NY, USA\\
Email: \href{mailto:ruy4006@med.cornell.edu}{ruy4006@med.cornell.edu}

\medskip
Chuan Hong, Department of Biostatistics and Bioinformatics, Duke University School of Medicine, Durham, NC, USA\\
Email: \href{mailto:chuan.hong@duke.edu}{chuan.hong@duke.edu}
\endgroup
\section*{Abstract}
{\AbstractContent}
\clearpage
\section{Introduction}
\input{sections/introduction}
\FloatBarrier
\section{Results}
\input{sections/results}
\FloatBarrier
\section{Discussion}
\input{sections/discussion}
\section{Methods}
\input{sections/methods}
\section*{Data Availability}
MIMIC-IV is available to credentialed users through PhysioNet.\cite{mimiciv} The Duke Health data contain protected health information and are not available for sharing.
\section*{Code Availability}
\textcolor{black}{Code will be made publicly available on GitHub upon publication.}
\section*{Ethics Statement}
Use of the Duke Health data was approved by the Duke University Health System Institutional Review Board (Pro00110219) with a waiver of informed consent. MIMIC-IV was accessed under the PhysioNet credentialed data use agreement.
\FloatBarrier
\setlength{\bibsep}{3pt plus 0.3ex}
\bibliographystyle{naturemag-hyphen}
\bibliography{references}
\clearpage
\setcounter{table}{0}
\setcounter{figure}{0}
\setcounter{equation}{0}
\renewcommand{\thetable}{S\arabic{table}}
\renewcommand{\thefigure}{S\arabic{figure}}
\renewcommand{\theequation}{S\arabic{equation}}
\renewcommand{\theHtable}{supp.\arabic{table}}
\renewcommand{\theHfigure}{supp.\arabic{figure}}
\renewcommand{\theHequation}{supp.\arabic{equation}}
\section*{Supplementary Information}
\input{sections/supplement_body}

\end{document}

%% file: preamble.tex
\documentclass[11pt,a4paper,nonumbering]{dm}
\usepackage{mathrsfs}
\usepackage[super,sort&compress]{natbib}
\usepackage{etoolbox}
\makeatletter
\patchcmd{\NAT@citexnum}
  {--\NAT@penalty}{-\NAT@penalty}
  {}{\PackageError{ehrdib}{natbib range patch failed}{Check the natbib citation-range definition.}}
\makeatother
\usepackage{multirow,array,longtable}
\usepackage{placeins,float,microtype}
\usepackage{xspace}
\usepackage{setspace}

\newcommand{\alttext}[1]{}
\setcitestyle{citesep={,}}
\DeclareCaptionLabelSeparator{naturepipe}{ $\vert$ }
\makeatletter
\renewcommand\AB@affilsepx{\protect\\[\affilsep]\protect\Affilfont}
\newlength{\authoraffilgap}
\renewcommand\@author{\ifx\AB@affillist\AB@empty\AB@author\else
  \ifnum\value{affil}>\value{Maxaffil}\def\rlap##1{##1}%
  \AB@authlist\\[\authoraffilgap]\AB@affillist
  \else \AB@authors\fi\fi}
\let\AbstractContent\abscontent
\renewcommand{\AbstractContent}{%
  \begingroup
  \noindent\parbox{\dimexpr\linewidth}{\absfont\theabstract\par}\par
  \endgroup
  \@ifundefined{@keywords}{}{\vskip1em\noindent\keywordsfont Keywords: \@keywords\par}}
\renewcommand{\abscontent}{}
\makeatother

%% file: title.tex
\title{\centering Evidence-Conditioned Completion and Synthesis of Electronic Health Records}
\author[1]{Sicheng Shu}
\author[2]{Parker Knight}
\author[3]{Huitao Li}
\author[3]{Yuan Luo}
\author[4,5]{Yifan Peng}
\author[3]{Shawn N. Murphy}
\author[4,5,\textdagger]{Rui~Yang}
\author[1,\textdagger]{Chuan~Hong}
\affil[1]{Department of Biostatistics and Bioinformatics, Duke University School of Medicine, Durham, NC, USA}
\affil[2]{Department of Biostatistics, University of North Carolina at Chapel Hill, Chapel Hill, NC, USA}
\affil[3]{Department of Biomedical Informatics and Medical Education, University of Washington, WA, USA}
\affil[4]{Department of Population Health Sciences, Weill Cornell Medicine, New York, NY, USA}
\affil[5]{Systems Engineering, Cornell University, Ithaca, NY, USA}

%% file: sections/abstract.tex
\begin{abstract}
Patient information can guide electronic health record (EHR) generation, but how its amount shapes predictive utility and disclosure risk remains poorly characterized. Here we present EHRDiB, which unifies visit completion and synthesis by distinguishing original missingness from withheld inputs and combining a shared multimodal representation with an evidence-conditioned latent diffusion bridge. We compared EHRDiB with three completion and eight synthesis methods on MIMIC-IV and numerical and discretized Duke Health representations. Evaluation covered downstream predictive utility, generative diversity, and distributional fidelity for both tasks, and disclosure risk for synthesis. EHRDiB achieved the smallest utility gaps from real-data-trained classifiers in seven of eight completion comparisons. In synthesis, AUPRC gaps were 45.2\% smaller for ten MIMIC-IV principal-diagnosis targets and 53.3\% smaller for numerical Duke Health data relative to the best comparator. Synthesis with calibrated thresholding yielded the lowest marginal distribution error across all three datasets, with Coverage above 0.95 for both output rules. Across twelve evidence-fraction intervals in MIMIC-IV synthesis, membership-inference AUROC increased from 0.516 to 0.598 as the mean target evidence fraction rose from 0.20 to 0.50; utility gaps narrowed relative to the lowest evidence configuration but did not decrease monotonically. Our study provides a unified framework and empirical foundation for EHR generation that considers both its utility for clinical analysis and its information disclosure risk.
\end{abstract}

%% file: sections/introduction.tex
Electronic health records (EHRs) capture rich relationships among diseases, treatments, and physiological measurements, but these records are often incomplete and difficult to share across institutions \cite{agniel2018biases,ehrbenchmark}. By learning clinical associations from existing records, generative models can complete partial patient records to support analysis of existing populations, or synthesize surrogate records for method development, validation, and data sharing. Patient information available to guide generation - including observed clinical findings, patterns of clinical information availability, and information from clinically related patient populations - can provide clinical context, but links between generated records and real individuals may pose disclosure risks \cite{stadler2022synthetic}. It is therefore necessary to examine how analytical utility and disclosure risk vary with the amount of clinical information used to guide generation.

Generative modeling can now handle heterogeneous clinical variables, preserve cross-modal and longitudinal relationships, and accommodate incomplete records and missing modalities \cite{yoon2023ehrsafe,ehrmgan,flexgenehr}. Conditional generative models can use observed clinical information to complete partially observed records or synthesize new records under specified clinical conditions \cite{vsae,wang2022promptehr}. Despite these advances, variational models face a trade-off between reconstruction accuracy and latent-space regularization \cite{tabsyn}. Mode collapse in adversarial models can limit coverage of diverse clinical feature combinations \cite{medgan}. Diffusion models operating directly on sparse visit-level records can struggle to recover clinical events and their cross-modal relationships after corruption \cite{ehrd3pm}. Beyond these modeling challenges, reproducing variable distributions well does not necessarily imply that generated records are diverse, that the relationships needed for analysis are well preserved, or that disclosure risk is lower \cite{ehrbenchmark,sella2025information}. These findings indicate that the value of generated data depends on their intended use, and choices about generation conditions therefore require evidence that reflects both utility and disclosure risk.

\begin{figure}[H]
\centering
\includegraphics[width=\linewidth,trim=3bp 0bp 76bp 0bp,clip]{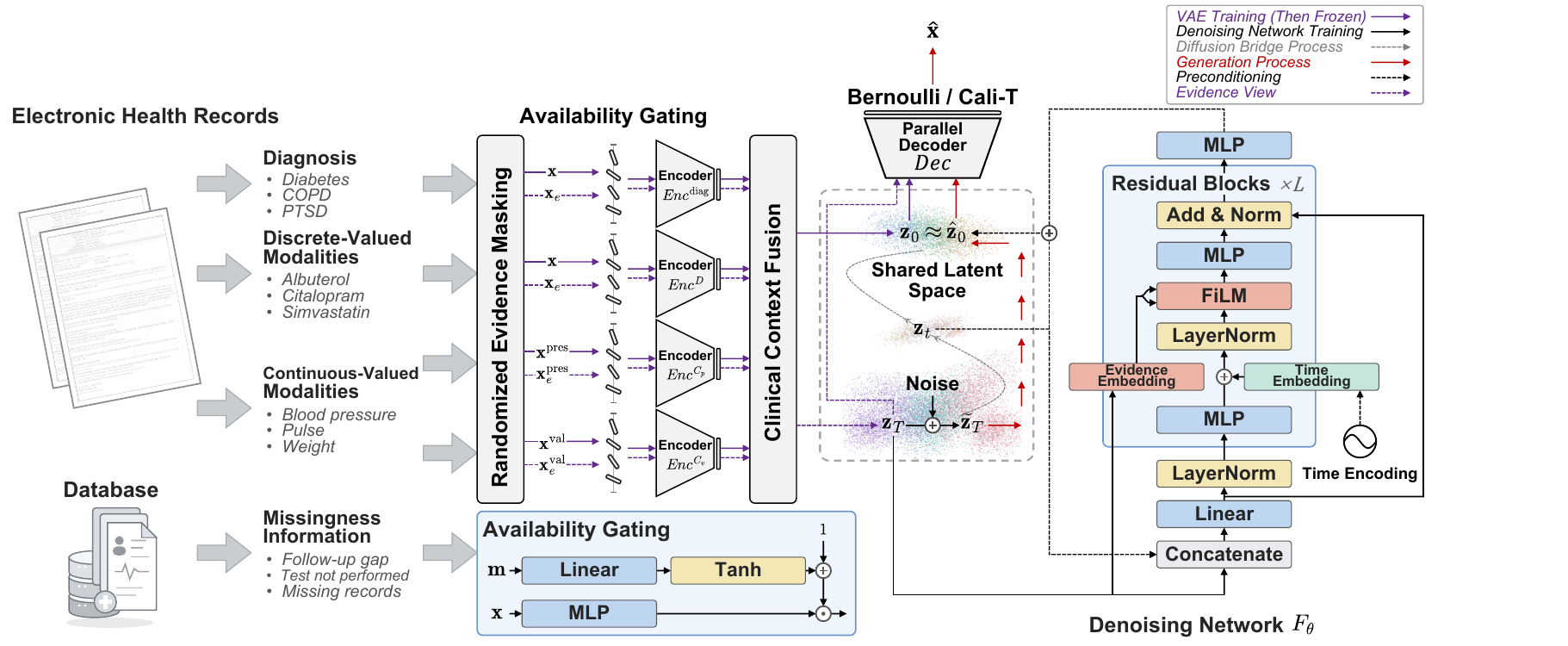}
\caption{Overview of EHRDiB. Availability-gated encoders and clinical context fusion map clinical information from data and evidence views into a shared latent space, with presence and values encoded separately for continuous-valued modalities. A latent diffusion bridge uses an evidence-conditioned denoising network to predict the data endpoint from intermediate latent states, guiding the generation of a visit representation from the evidence representation. The starting distribution is perturbed for synthesis. Parallel decoders map the generated representation to all clinical modalities, with two output rules: Bernoulli sampling and calibrated thresholding (Cali-T). Visual elements adapted from Li et al.~\cite{ehrmgan}; example EHR image from VA Blue Button~\cite{vabluebutton}.}
\label{fig:workflow}
\alttext{Schematic of the EHRDiB framework. Clinical modalities of a visit pass through availability-gated encoders into a shared latent space, where a diffusion bridge connects an evidence representation to a full-visit representation; parallel decoders produce the generated visit.}
\end{figure}

In this study, we introduce EHRDiB, an \textbf{EHR} \textbf{Di}ffusion \textbf{B}ridge framework that unifies the completion of partially observed visit records and the synthesis of new records while systematically varying the amount of supplied clinical information to assess the utility gains and disclosure risks associated with additional patient information (\textbf{Fig.~\ref{fig:workflow}}). EHRDiB models heterogeneous clinical variables using type-appropriate distributions, incorporates clinical information availability into its representations, explicitly identifies which clinical information is supplied as generation conditions, and models measurement values separately from their presence in the original record. These representations are integrated through clinical context fusion (CCF) guided by the principal diagnosis and coupled with an evidence-conditioned latent diffusion bridge.

We compared EHRDiB with three completion baselines and eight synthesis baselines on MIMIC-IV and Duke Health data in numerical (DH) and discretized (DH-D) representations. Evaluation covered downstream predictive utility, generative diversity, and distributional fidelity for both tasks, and disclosure risk for synthesis. EHRDiB achieved the smallest utility gaps from real-data-trained classifiers in seven of eight completion comparisons. In synthesis, EHRDiB with Bernoulli sampling reduced the AUPRC gap relative to real-data training by 45.2\% on the ten MIMIC-IV principal-diagnosis prediction tasks and by 53.3\% on numerical Duke Health data, compared with the best-performing baseline in each setting. With calibrated thresholding (Cali-T), EHRDiB achieved the lowest marginal distribution error across all three data settings. Both output rules maintained Coverage above 0.95, indicating broad coverage of real-data neighborhoods. Across twelve MIMIC-IV synthesis configurations, increasing the mean proportion of supplied clinical variables from 0.20 to 0.50 was associated with membership-inference AUROC rising from 0.516 to 0.598. Utility gaps narrowed relative to the lowest evidence configuration but did not decrease monotonically. Our study provides a unified framework and empirical evidence for assessing downstream predictive utility and disclosure risk under different clinical conditioning settings, while maintaining high distributional fidelity and generative diversity.

%% file: sections/results.tex
\label{sec:results}

\subsection{Study Population}
\label{sec:results-population}

The MIMIC-IV cohort comprised 245,130 visits from 119,425 patients. The estimated median age at the first included admission was 61 years (IQR 43-74), and 52.6\% were female. The Duke Health cohort comprised 936,471 visits from 120,856 patients seen between 2015 and 2018. We analyzed the same Duke Health visits in two representations: Duke Health (DH), which retained clinical measurements as numerical values, and Duke Health Discrete (DH-D), which discretized these values into categories using binning. The mean age at the PREVENT study baseline was 53.7 years (SD 12.5), and 56.8\% were female. A median of 7.8\% of clinical measurement variables was present per MIMIC-IV visit and 25\% per Duke visit. Prevalence of the ten principal-diagnosis targets ranged from 0.5\% to 3.5\% in MIMIC-IV and from 1.8\% to 27.9\% in the Duke data. Detailed information on the study populations is provided in \textbf{Supplementary Tab.~\ref{tab:population}}.

\subsection{Completion}
\label{sec:results-completion}

We first evaluated the predictive utility of completed records for phenotyping in MIMIC-IV and Duke Health. MIMIC-IV included two downstream tasks targeting ten principal diagnoses and seven Phecode-defined phenotypes. For Duke Health, we evaluated the same ten diagnosis targets in DH and DH-D. Relative to LightGBM classifiers trained on real data, EHRDiB achieved the smallest performance gaps in seven of eight comparisons across the four phenotyping tasks and two metrics, AUROC and AUPRC (\textbf{Tab.~\ref{tab:completion}}). For the ten MIMIC-IV principal-diagnosis targets, EHRDiB's \(\Delta\)AUROC and \(\Delta\)AUPRC were 0.0021 and 0.0319, representing reductions of 23.0\% and 19.5\% relative to the best comparators for the respective metrics, GAIN and CSDI. In DH and DH-D, its \(\Delta\)AUPRC values were 0.0152 and 0.0150, representing reductions of 38.5\% and 42.6\% relative to the best comparators in the respective datasets, CISS-VAE and GAIN.

Beyond predictive utility, we examined the diversity of completed records using Coverage and the average pairwise distance ratio (APDR), as defined in \hyperref[sec:evaluation]{Methods}. EHRDiB achieved the highest Coverage in MIMIC-IV, DH, and DH-D, with values of 0.9938, 0.9708, and 0.9627, respectively, indicating broad coverage of real-data neighborhoods. EHRDiB yielded APDR values of 0.9434, 0.9044, and 0.9357 in MIMIC-IV, DH, and DH-D, respectively. Its APDR performance was slightly below that of CSDI across all three datasets, suggesting more conservative completion.

For distributional fidelity, CSDI achieved the lowest feature distribution distance (FDD) in all three datasets and the lowest blockwise correlation matrix distance (BCMD) in DH and DH-D. However, EHRDiB achieved the lowest BCMD in MIMIC-IV at 11.93, compared with 12.85 for CSDI, indicating closer agreement with real-data correlations.

\input{tables/completion}

\subsection{Synthesis}
\label{sec:results-synthesis}

\begin{figure}[H]
\centering
\includegraphics[width=.97\linewidth]{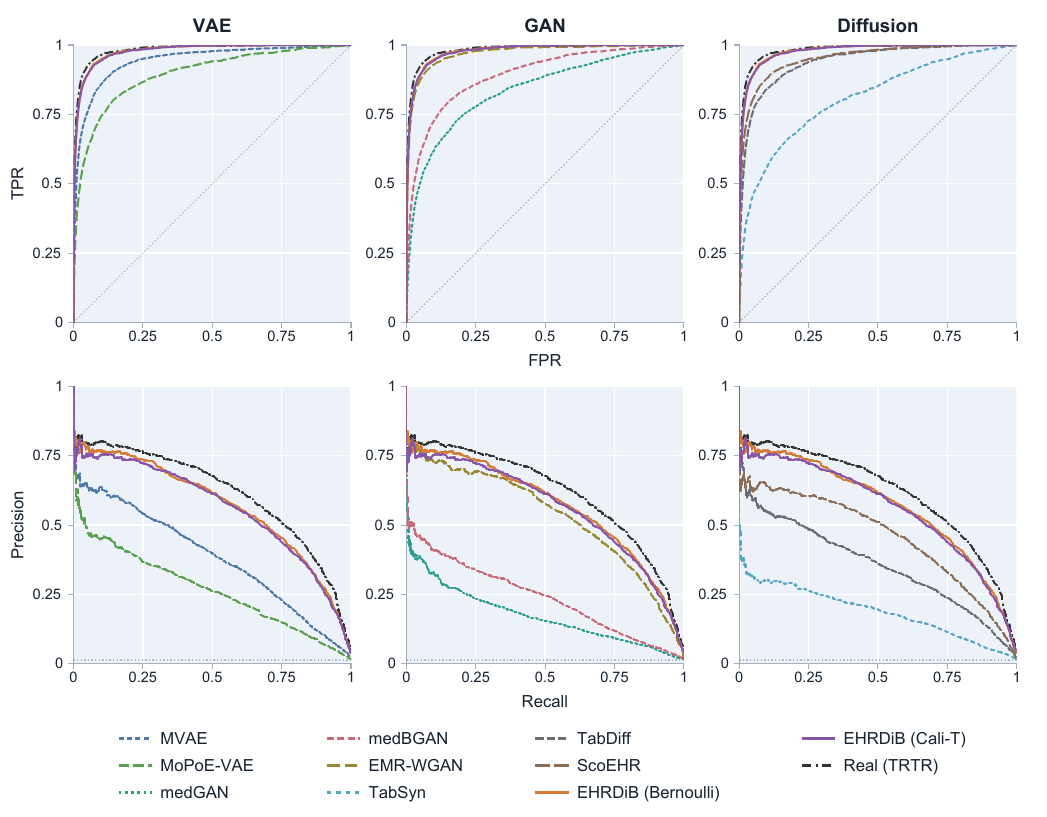}
\caption{MIMIC-IV synthesis utility for the ten principal-diagnosis targets. Macro-averaged ROC (top) and precision-recall (bottom) curves compare LightGBM classifiers trained on real (TRTR) or synthetic records and evaluated on the same real test set. Columns group variational, adversarial, and diffusion comparators, with EHRDiB (Bernoulli) and EHRDiB (Cali-T) shown in each.}
\label{fig:utility}
\alttext{Six panels of receiver operating characteristic and precision-recall curves for classifiers trained on synthetic MIMIC-IV visits from each method, compared with a classifier trained on real visits.}
\end{figure}

We next evaluated the predictive utility of synthetic records for the same phenotyping tasks. Relative to LightGBM classifiers trained on real data, EHRDiB (Bernoulli) achieved the smallest \(\Delta\)AUROC and \(\Delta\)AUPRC for the ten MIMIC-IV principal-diagnosis targets, at 0.0042 and 0.0457, representing reductions of 63.4\% and 45.2\% relative to EMR-WGAN, the best comparator for both metrics (\textbf{Tab.~\ref{tab:synthesis}; Fig.~\ref{fig:utility}}). For the seven Phecode-defined phenotypes, EHRDiB (Cali-T) achieved the smallest gaps at 0.0388 and 0.0947, compared with 0.0411 and 0.0962 for EMR-WGAN (\textbf{Supplementary Fig.~\ref{fig:utility-b}}). Target-specific ROC curves are provided in \textbf{Supplementary Figs.~\ref{fig:diagnosis-a} and~\ref{fig:diagnosis-b}}.

In DH, EHRDiB (Bernoulli) achieved the smallest \(\Delta\)AUROC and \(\Delta\)AUPRC at 0.0055 and 0.0149, representing reductions of 71.4\% and 53.3\% relative to EMR-WGAN, the best comparator. In DH-D, its \(\Delta\)AUROC and \(\Delta\)AUPRC were 0.0059 and 0.0136, respectively. Its \(\Delta\)AUPRC was the smallest among all models, compared with 0.0140 for ScoEHR, whereas its \(\Delta\)AUROC was higher than ScoEHR's 0.0042.

Beyond predictive utility, we also used Coverage and APDR to assess the diversity of synthetic records. In MIMIC-IV, EHRDiB (Cali-T) achieved the highest Coverage at 0.9825, a gain of 17.1 percentage points over MVAE (0.8116), the best comparator. EHRDiB (Bernoulli) achieved the highest Coverage in DH and DH-D at 0.9631 and 0.9648, respectively, exceeding the best comparators, EMR-WGAN (0.8687) and ScoEHR (0.9510), by 9.4 and 1.4 percentage points. EHRDiB (Cali-T) achieved APDR closest to one in MIMIC-IV and DH, with values of 0.9683 and 0.9680. In DH-D, its APDR was 1.0039, slightly farther from one than medBGAN's 1.0036.

\input{tables/synthesis}

\subsection{Marginal Distribution and Modality Correlation}
\label{sec:results-fidelity}

\begin{figure}[H]
\centering
\includegraphics[width=\linewidth]{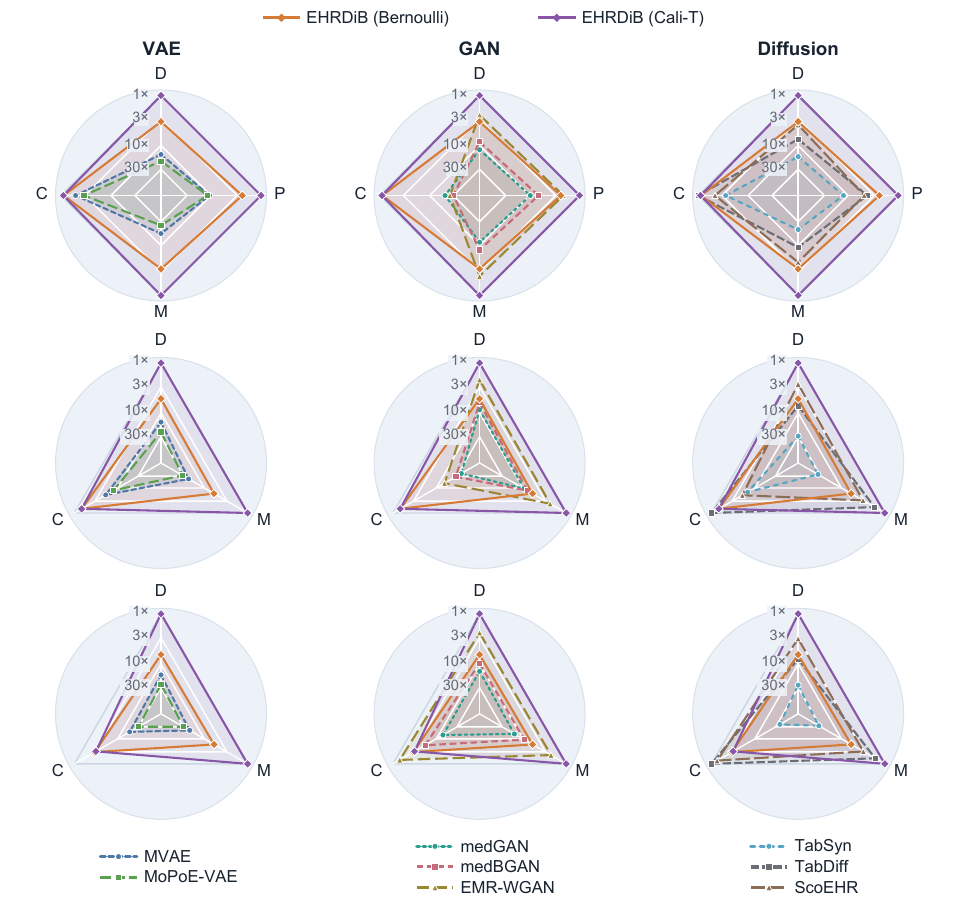}
\caption{Marginal distribution fidelity of synthetic records in MIMIC-IV (top), DH (middle), and DH-D (bottom). D, diagnosis; P, procedure (MIMIC-IV only); M, medication; C, clinical measurement. Each modality's FDD is normalized to the lowest value among models within the same dataset. The reversed logarithmic radial scale runs from \(1\times\) at the outer ring to \(100\times\) at the center; larger radii indicate lower errors. Columns show variational, adversarial, and diffusion baselines alongside EHRDiB under both output rules.}
\label{fig:fdd}
\end{figure}

We then assessed how well synthetic records preserved marginal distributions. EHRDiB (Cali-T) achieved the lowest FDD in all three datasets (\textbf{Tab.~\ref{tab:synthesis}}). In MIMIC-IV, its FDD was 21.10, 3.4\% lower than TabDiff's 21.84. In DH, its FDD was 0.78, 52.8\% lower than TabDiff's 1.66, representing the largest relative improvement. In DH-D, FDD was 0.45, 20.0\% lower than EMR-WGAN's 0.56. The modality-level comparison showed that EHRDiB (Cali-T) achieved the lowest FDD for diagnoses and medications across all three datasets, and for procedures in MIMIC-IV. Its advantage over EHRDiB (Bernoulli) was most pronounced in these clinical code modalities, whereas TabDiff achieved the lowest FDD for clinical measurements (\textbf{Fig.~\ref{fig:fdd}}).

\begin{figure}[H]
\centering
\includegraphics[width=\linewidth]{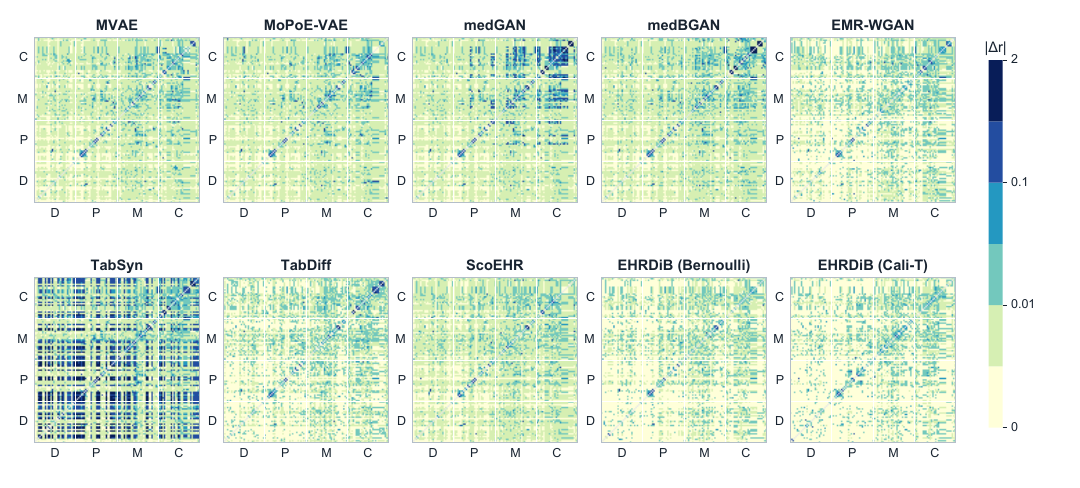}
\caption{Absolute Pearson-correlation errors between real and synthetic MIMIC-IV records. All models share the same 120 features, with 30 each from diagnosis (D), procedure (P), medication (M), and clinical measurement (C). Darker colors indicate larger errors on a shared scale. This exploratory subset was selected using test-set error contrasts.}
\label{fig:bcmd}
\end{figure}

Beyond marginal distributions, we evaluated the preservation of within- and cross-modality correlations. EHRDiB (Bernoulli) achieved the lowest BCMD in all three datasets (\textbf{Tab.~\ref{tab:synthesis}}). In MIMIC-IV, BCMD was 14.72, a 14.1\% reduction from EMR-WGAN's 17.14. In DH, it was 12.05, 50.1\% lower than EMR-WGAN's 24.15. In DH-D, it was 8.40, 4.5\% lower than EMR-WGAN's 8.79. In the displayed 120-feature MIMIC-IV subset, EHRDiB under both output rules showed lower errors within the clinical-measurement block than the comparators. EHRDiB (Bernoulli) further reduced errors within procedures and between procedures and clinical measurements relative to EHRDiB (Cali-T) (\textbf{Fig.~\ref{fig:bcmd}}). EHRDiB showed complementary strengths under the two output rules across all three datasets: EHRDiB (Cali-T) achieved lower FDD, whereas EHRDiB (Bernoulli) achieved lower BCMD.

\subsection{Disclosure Risk and Supplied Evidence}
\label{sec:results-evidence}

We next evaluated synthesis disclosure risk using a membership inference attack (MIA) and an attribute inference attack (AIA) (\textbf{Tab.~\ref{tab:synthesis-privacy}}). Across the two output rules for EHRDiB, MIA AUROC ranged from 0.5336 to 0.5393 in MIMIC-IV, 0.5105 to 0.5119 in DH, and 0.5138 to 0.5162 in DH-D. These values exceeded the range across comparator models in the three datasets (0.4920-0.5078), indicating a stronger membership-inference signal under this attack. For AIA, EHRDiB (Bernoulli) had negative attribute-inference advantages (\(\Delta\)AIA) of \(-0.0049\), \(-0.0065\), and \(-0.0046\) in MIMIC-IV, DH, and DH-D, respectively. Negative values indicate lower attack AUPRC when synthetic records were used as the reference compared with real auxiliary data. EHRDiB (Cali-T) also had negative \(\Delta\)AIA in DH and DH-D, but a small positive value in MIMIC-IV (0.0035). All comparators had negative \(\Delta\)AIA.

\input{tables/privacy}

\begin{figure}[H]
\centering
\includegraphics[width=\linewidth]{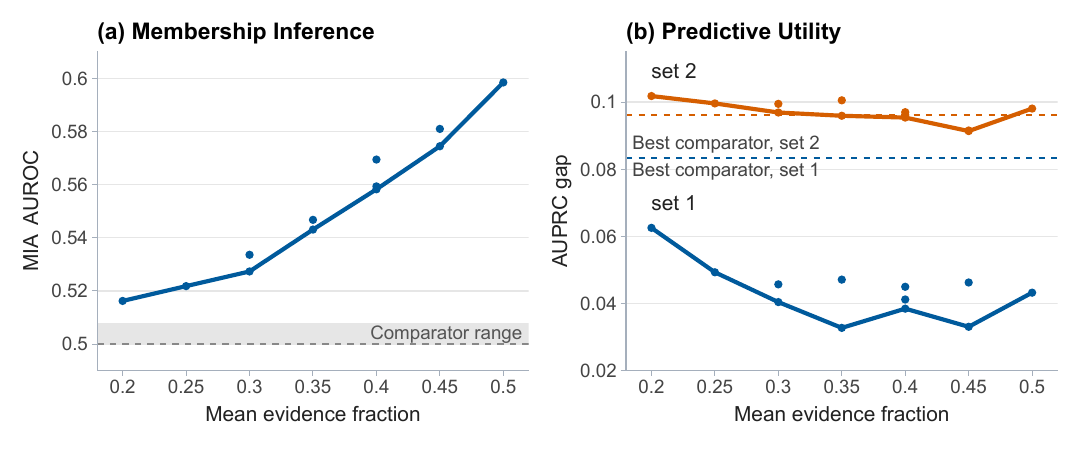}
\caption{Disclosure risk and phenotyping utility in MIMIC-IV synthesis across evidence-fraction intervals. Points show individual intervals at their mean evidence fraction; each line connects the minimum observed value at each mean. (a) MIA AUROC; the gray band shows the comparator range and the dashed line marks random discrimination (0.5). (b) AUPRC gaps for set 1 (ten principal-diagnosis targets) and set 2 (seven Phecode-defined phenotypes); dashed lines indicate the best comparator for each set.}
\label{fig:evidence-axis}
\alttext{Two panels show EHRDiB results across twelve evidence-fraction intervals, with points at mean evidence fractions from 0.20 to 0.50. Panel a shows MIA AUROC rising from about 0.52 to 0.60, above a gray band around 0.50 representing the comparator range. Panel b shows nonmonotonic AUPRC gaps for two phenotyping tasks, with dashed lines marking the best comparator values. Each line connects the minimum observed value of its metric at each mean.}
\end{figure}

To examine how disclosure risk varied with the amount of input information, we evaluated EHRDiB across twelve evidence-fraction intervals in MIMIC-IV synthesis. For each reference visit, we drew one target evidence fraction from a uniform distribution over the specified interval. Within each modeling group, this fraction determined the number of eligible clinical variables randomly selected without replacement and supplied as generation conditions (\hyperref[sec:formulation]{Methods}). MIA AUROC increased from 0.5162 to 0.5984 as the mean evidence fraction rose from 0.20 to 0.50, indicating that more input information was associated with a stronger membership-inference signal (\textbf{Fig.~\ref{fig:evidence-axis}a}). At the same mean fraction of 0.40, the intervals \([0.1,0.7]\), \([0.2,0.6]\), and \([0.3,0.5]\) yielded MIA AUROCs of 0.5694, 0.5593, and 0.5582, respectively. \(\Delta\)AIA remained negative across all twelve intervals, ranging from \(-0.0049\) to \(-0.0023\).

Even at the lowest evidence-fraction interval, EHRDiB's utility gaps for the ten principal-diagnosis targets remained smaller than those of every comparator. Relative to the lowest evidence configuration, all four utility gaps narrowed in the other eleven configurations, but increasing the amount of input information did not yield monotonic improvements in phenotyping utility (\textbf{Fig.~\ref{fig:evidence-axis}b; Supplementary Tab.~\ref{tab:robustness-utility}}). For the ten principal-diagnosis targets, \(\Delta\)AUROC and \(\Delta\)AUPRC ranged from 0.0027 to 0.0054 and from 0.0327 to 0.0625, respectively. For the seven Phecode-defined phenotypes, the corresponding ranges were 0.0342-0.0426 and 0.0914-0.1018. The smallest \(\Delta\)AUPRC values for the two tasks occurred at \([0.2,0.5]\) and \([0.2,0.7]\), respectively, whereas the highest evidence configuration, \([0.3,0.7]\), yielded values of 0.0432 and 0.0981, both above their respective minima.

Across the twelve evidence-fraction intervals, EHRDiB maintained Coverage of 0.9759-0.9933. From the lowest to the highest evidence configuration, APDR increased from 0.9407 to 0.9632, moving closer to one, while FDD decreased from 24.01 to 21.19 and BCMD from 15.43 to 13.70. The lowest FDD (20.84) and the APDR closest to one (0.9724) both occurred at \([0.3,0.6]\), indicating that the highest evidence configuration did not achieve the best result for every measure. Full configuration-level results are provided in \textbf{Supplementary Tabs.~\ref{tab:robustness-fidelity}-\ref{tab:robustness-privacy}}.

\subsection{Model Components and Ablation Study}
\label{sec:results-components}

\textcolor{black}{We finally examined the contributions of model components in MIMIC-IV synthesis. We first compared four approaches to encoding clinical information availability. Compared with no explicit availability encoding, gating reduced FDD from 23.11 to 22.11 and BCMD from 15.19 to 14.94. Coverage increased from 0.986 to 0.988, while the \(\Delta\)AUROC and \(\Delta\)AUPRC gaps decreased from 0.0064 to 0.0058 and from 0.0693 to 0.0595, respectively. Gating achieved the best values for these five measures among the four approaches. Availability concatenation had the APDR closest to one (0.9582), slightly higher than gating's 0.9576 (\textbf{Supplementary Tab.~\ref{tab:availability}}).}

\textcolor{black}{Kullback-Leibler (KL) regularization strength and forward bridge choice further influenced generation performance. Across KL weights from \(10^{-6}\) to \(10^{-1}\), \(10^{-5}\) achieved the lowest FDD and BCMD (22.11 and 14.94), APDR closest to one (0.9576), and the smallest \(\Delta\)AUROC and \(\Delta\)AUPRC (0.0058 and 0.0595). The two largest weights, \(10^{-2}\) and \(10^{-1}\), degraded fidelity and phenotyping utility. Coverage was highest at \(10^{-4}\) (0.9891, compared with 0.9884 at \(10^{-5}\)) (\textbf{Supplementary Tab.~\ref{tab:kl}}). We also compared VE, VP, Linear, and Schr\"odinger forward bridge paths. The Schr\"odinger path achieved the lowest FDD and BCMD (20.91 and 14.37), APDR closest to one (0.9741), and the smallest \(\Delta\)AUROC (0.0047). Linear yielded slightly higher Coverage (0.9896 versus 0.9892) and a slightly smaller \(\Delta\)AUPRC (0.0514 versus 0.0517) than the Schr\"odinger path (\textbf{Supplementary Tab.~\ref{tab:bridge}}).}

\textcolor{black}{Finally, we evaluated how evidence-endpoint perturbation and sampling stochasticity affected EHRDiB synthesis. For the hybrid sampler, an unperturbed endpoint with a stochasticity level of 0.125 yielded the lowest FDD (20.85), whereas perturbation of 1 with zero stochasticity yielded the smallest \(\Delta\)AUPRC (0.0458). For the stochastic-interpolant sampler, the lowest FDD (21.33) occurred at perturbation 0.5 and stochasticity 1, whereas the smallest \(\Delta\)AUPRC (0.0464) occurred at perturbation 1 and stochasticity 0.5. Neither sampler had a configuration that led all measures within its tested grid. Sampling hyperparameters should therefore be selected according to the specific experimental configuration and the priorities given to fidelity, diversity, and phenotyping utility (\textbf{Supplementary Tabs.~\ref{tab:sam-hybrid} and~\ref{tab:sam-interpolant}}).}

%% file: tables/completion.tex
\begin{table}[H]
\centering
\caption{Completion results across three datasets. Utility gaps are differences in AUROC and AUPRC between LightGBM classifiers trained on real and completed records. Set 1 comprises ten principal-diagnosis targets in MIMIC-IV and ten diagnosis targets in DH and DH-D; set 2 comprises seven Phecode-defined phenotypes in MIMIC-IV. Bold indicates the best result for each metric within each dataset.}
\label{tab:completion}
\begingroup\fontsize{9}{11}\selectfont
\setlength{\tabcolsep}{4pt}\renewcommand{\arraystretch}{1.05}
\begin{tabular}{@{}lrrrrrrrr@{}}
\toprule
 & \multicolumn{2}{c}{Utility, set 1} & \multicolumn{2}{c}{Utility, set 2} & \multicolumn{2}{c}{Diversity} & \multicolumn{2}{c}{Fidelity (\(\times10^{3}\))} \\
\cmidrule(lr){2-3}\cmidrule(lr){4-5}\cmidrule(lr){6-7}\cmidrule(l){8-9}
Model & \(\Delta\)AUROC & \(\Delta\)AUPRC & \(\Delta\)AUROC & \(\Delta\)AUPRC & Coverage & APDR & FDD & BCMD \\
\midrule
\multicolumn{9}{@{}l}{\textit{MIMIC-IV}} \\
CISS-VAE & 0.0048 & 0.0624 & 0.0334 & 0.0848 & 0.9417 & 0.8296 & 16.25 & 13.34 \\
GAIN & 0.0027 & 0.0504 & 0.0316 & \textbf{0.0787} & 0.6076 & 0.9257 & 21.69 & 16.76 \\
CSDI & 0.0038 & 0.0396 & 0.0396 & 0.0873 & 0.9711 & \textbf{0.9671} & \textbf{9.70} & 12.85 \\
EHRDiB & \textbf{0.0021} & \textbf{0.0319} & \textbf{0.0314} & 0.0837 & \textbf{0.9938} & 0.9434 & 13.17 & \textbf{11.93} \\
\midrule
\multicolumn{9}{@{}l}{\textit{DH}} \\
CISS-VAE & 0.0094 & 0.0248 & - & - & 0.7443 & 0.8958 & 3.60 & 15.95 \\
GAIN & 0.0095 & 0.0259 & - & - & 0.7819 & 0.9277 & 4.76 & 26.26 \\
CSDI & 0.0108 & 0.0287 & - & - & 0.9280 & \textbf{1.0135} & \textbf{0.97} & \textbf{13.24} \\
EHRDiB & \textbf{0.0060} & \textbf{0.0152} & - & - & \textbf{0.9708} & 0.9044 & 1.80 & 15.32 \\
\midrule
\multicolumn{9}{@{}l}{\textit{DH-D}} \\
CISS-VAE & 0.0206 & 0.0594 & - & - & 0.8674 & 0.9116 & 3.33 & 9.69 \\
GAIN & 0.0103 & 0.0262 & - & - & 0.8014 & 0.8431 & 3.88 & 10.34 \\
CSDI & 0.0130 & 0.0285 & - & - & 0.9257 & \textbf{0.9697} & \textbf{0.61} & \textbf{7.86} \\
EHRDiB & \textbf{0.0059} & \textbf{0.0150} & - & - & \textbf{0.9627} & 0.9357 & 1.44 & 9.86 \\
\bottomrule
\end{tabular}
\endgroup
\end{table}

%% file: tables/synthesis.tex
\begin{table}[H]
\centering
\caption{Synthesis results across three datasets. Utility gaps are differences in AUROC and AUPRC between LightGBM classifiers trained on real and synthetic records. Set 1 comprises ten principal-diagnosis targets in MIMIC-IV and ten diagnosis targets in DH and DH-D; set 2 comprises seven Phecode-defined phenotypes in MIMIC-IV. Bold indicates the best result for each metric within each dataset.}
\label{tab:synthesis}
\begingroup\fontsize{9}{11}\selectfont
\setlength{\tabcolsep}{4pt}\renewcommand{\arraystretch}{1.05}
\begin{tabular}{@{}lrrrrrrrr@{}}
\toprule
 & \multicolumn{2}{c}{Utility, set 1} & \multicolumn{2}{c}{Utility, set 2} & \multicolumn{2}{c}{Diversity} & \multicolumn{2}{c}{Fidelity (\(\times10^{3}\))} \\
\cmidrule(lr){2-3}\cmidrule(lr){4-5}\cmidrule(lr){6-7}\cmidrule(l){8-9}
Model & \(\Delta\)AUROC & \(\Delta\)AUPRC & \(\Delta\)AUROC & \(\Delta\)AUPRC & Coverage & APDR & FDD & BCMD \\
\midrule
\multicolumn{9}{@{}l}{\textit{MIMIC-IV}} \\
MVAE & 0.0381 & 0.2310 & 0.1272 & 0.1487 & 0.8116 & 0.7442 & 42.89 & 25.52 \\
MoPoE-VAE & 0.0876 & 0.3479 & 0.1913 & 0.2165 & 0.5956 & 0.6548 & 62.38 & 32.08 \\
medGAN & 0.1440 & 0.4412 & 0.1831 & 0.2386 & 0.0663 & 4.8867 & 382.79 & 38.86 \\
medBGAN & 0.0897 & 0.3755 & 0.1474 & 0.2126 & 0.0890 & 5.0766 & 556.94 & 33.57 \\
EMR-WGAN & 0.0116 & 0.0835 & 0.0411 & 0.0962 & 0.1673 & 2.5711 & 474.64 & 17.14 \\
TabSyn & 0.1722 & 0.4261 & 0.1306 & 0.1624 & 0.7012 & 0.8381 & 73.33 & 203.72 \\
TabDiff & 0.0406 & 0.2586 & 0.0510 & 0.1331 & 0.5736 & 1.0382 & 21.84 & 18.47 \\
ScoEHR & 0.0291 & 0.1587 & 0.0793 & 0.1441 & 0.8083 & 0.7768 & 42.02 & 20.47 \\
EHRDiB (Bernoulli) & \textbf{0.0042} & \textbf{0.0457} & 0.0410 & 0.0969 & 0.9815 & 0.9548 & 22.38 & \textbf{14.72} \\
EHRDiB (Cali-T) & 0.0049 & 0.0551 & \textbf{0.0388} & \textbf{0.0947} & \textbf{0.9825} & \textbf{0.9683} & \textbf{21.10} & 16.63 \\
\midrule
\multicolumn{9}{@{}l}{\textit{DH}} \\
MVAE & 0.1240 & 0.1922 & - & - & 0.4954 & 0.7220 & 6.69 & 37.84 \\
MoPoE-VAE & 0.3098 & 0.3173 & - & - & 0.0392 & 0.4800 & 9.97 & 61.81 \\
medGAN & 0.1015 & 0.1617 & - & - & 0.4421 & 1.6418 & 15.79 & 191.86 \\
medBGAN & 0.0610 & 0.1090 & - & - & 0.6712 & 1.3563 & 11.83 & 43.96 \\
EMR-WGAN & 0.0193 & 0.0318 & - & - & 0.8687 & 1.0703 & 6.09 & 24.15 \\
TabSyn & 0.3967 & 0.3730 & - & - & 0.0974 & 0.3662 & 10.65 & 64.66 \\
TabDiff & 0.3758 & 0.3606 & - & - & 0.3216 & 0.9549 & 1.66 & 48.93 \\
ScoEHR & 0.1552 & 0.2091 & - & - & 0.5017 & 0.8372 & 2.57 & 32.98 \\
EHRDiB (Bernoulli) & \textbf{0.0055} & \textbf{0.0149} & - & - & \textbf{0.9631} & 0.9352 & 1.97 & \textbf{12.05} \\
EHRDiB (Cali-T) & 0.0069 & 0.0163 & - & - & 0.9580 & \textbf{0.9680} & \textbf{0.78} & 27.27 \\
\midrule
\multicolumn{9}{@{}l}{\textit{DH-D}} \\
MVAE & 0.0875 & 0.1274 & - & - & 0.5944 & 0.9022 & 5.15 & 21.42 \\
MoPoE-VAE & 0.3070 & 0.3370 & - & - & 0.0399 & 0.6277 & 7.78 & 33.83 \\
medGAN & 0.1050 & 0.1678 & - & - & 0.6482 & 0.9473 & 3.98 & 27.68 \\
medBGAN & 0.0669 & 0.1092 & - & - & 0.7493 & \textbf{1.0036} & 2.28 & 21.90 \\
EMR-WGAN & 0.0188 & 0.0346 & - & - & 0.9262 & 1.0039 & 0.56 & 8.79 \\
TabSyn & 0.3862 & 0.3726 & - & - & 0.0765 & 0.4801 & 8.73 & 30.15 \\
TabDiff & 0.4137 & 0.3659 & - & - & 0.3996 & 0.9374 & 1.07 & 18.63 \\
ScoEHR & \textbf{0.0042} & 0.0140 & - & - & 0.9510 & 0.9949 & 0.71 & 9.98 \\
EHRDiB (Bernoulli) & 0.0059 & \textbf{0.0136} & - & - & \textbf{0.9648} & 0.9516 & 1.44 & \textbf{8.40} \\
EHRDiB (Cali-T) & 0.0066 & 0.0139 & - & - & 0.9610 & 1.0039 & \textbf{0.45} & 15.23 \\
\bottomrule
\end{tabular}
\endgroup
\end{table}

%% file: tables/privacy.tex
\begin{table}[H]
\centering
\caption{Synthesis disclosure risk across three datasets. Membership inference attack (MIA) performance is measured by AUROC, with chance at 0.5. Attribute inference advantage (\(\Delta\)AIA) is the difference in attack AUPRC between a synthetic reference and a real auxiliary reference. Bold indicates MIA AUROC closest to 0.5 and the lowest \(\Delta\)AIA within each dataset.}
\label{tab:synthesis-privacy}
\begingroup\fontsize{9}{11}\selectfont
\setlength{\tabcolsep}{3pt}\renewcommand{\arraystretch}{1.05}
\begin{tabular}{@{}l@{\hspace{12pt}}r@{\hspace{6pt}}r@{\hspace{10pt}}r@{\hspace{6pt}}r@{\hspace{10pt}}r@{\hspace{6pt}}r@{}}
\toprule
 & \multicolumn{2}{c}{MIMIC-IV} & \multicolumn{2}{c}{DH} & \multicolumn{2}{c}{DH-D} \\
Model & \multicolumn{1}{c}{MIA} & \multicolumn{1}{c}{\(\Delta\)AIA} & \multicolumn{1}{c}{MIA} & \multicolumn{1}{c}{\(\Delta\)AIA} & \multicolumn{1}{c}{MIA} & \multicolumn{1}{c}{\(\Delta\)AIA} \\
\midrule
MVAE & 0.5068 & -0.0161 & 0.4983 & -0.0379 & 0.4990 & -0.0291 \\
MoPoE-VAE & 0.5010 & -0.0325 & 0.4969 & -0.0826 & 0.4920 & -0.0774 \\
medGAN & 0.4995 & -0.0358 & 0.5017 & -0.0414 & 0.5018 & -0.0511 \\
medBGAN & \textbf{0.5002} & \textbf{-0.0546} & \textbf{0.5013} & -0.0289 & 0.5013 & -0.0370 \\
EMR-WGAN & 0.5078 & -0.0103 & 0.5028 & -0.0062 & 0.5031 & -0.0095 \\
TabSyn & 0.5023 & -0.0490 & 0.5023 & \textbf{-0.1229} & 0.5050 & \textbf{-0.1169} \\
TabDiff & 0.5039 & -0.0270 & 0.5022 & -0.1224 & 0.5037 & -0.1161 \\
ScoEHR & 0.5074 & -0.0061 & 0.5046 & -0.0078 & \textbf{0.5003} & -0.0061 \\
EHRDiB (Bernoulli) & 0.5336 & -0.0049 & 0.5119 & -0.0065 & 0.5162 & -0.0046 \\
EHRDiB (Cali-T) & 0.5393 & 0.0035 & 0.5105 & -0.0031 & 0.5138 & -0.0026 \\
\bottomrule
\end{tabular}
\endgroup
\end{table}

%% file: sections/discussion.tex
\label{sec:discussion}

This study evaluated EHRDiB, a unified framework for evidence-conditioned EHR completion and synthesis. Results from MIMIC-IV and Duke Health showed that the framework supported downstream tasks and covered real-data neighborhoods in both tasks. The MIMIC-IV evidence-fraction analysis further indicated that supplying more patient information was associated with a stronger membership-inference signal, while predictive utility did not improve monotonically.

EHR completion and synthesis should generate records that support downstream tasks while maintaining good distributional fidelity and diversity. In completion, although CSDI more closely matched the marginal distributions and overall dispersion of real data, EHRDiB achieved smaller utility gaps in all four phenotyping tasks and broader coverage of real-data neighborhoods across all three datasets. In synthesis, EHRDiB (Bernoulli) achieved smaller utility gaps in three of the four phenotyping tasks and better preserved correlations, whereas EHRDiB (Cali-T) more closely matched marginal distributions and overall dispersion. Cali-T calibrates output thresholds using training-set prevalence, so part of its marginal fidelity advantage is expected by design.

EHRDiB showed clear advantages over the baselines in predictive utility, diversity, and distributional fidelity on the mixed-type MIMIC-IV and DH datasets, and remained among the leading methods on DH-D. Diffusion baselines did not consistently outperform GAN or VAE baselines across datasets and representations, while the evaluated VAE baselines showed severe shortcomings in downstream utility and coverage. These limitations further motivate EHRDiB's integration of variational representation learning with an evidence-conditioned latent diffusion bridge.

Across the evidence-fraction intervals evaluated in MIMIC-IV synthesis, utility improved relative to the lowest evidence configuration, but further increases in supplied information did not yield monotonic gains and were associated with a stronger membership-inference signal. These findings support tailoring the evidence fraction to the intended use of generated data. Completion retains supplied patient information and is suited to analyses within the environment holding the source records. For synthesis, evidence fractions should be chosen to meet downstream utility requirements while accounting for disclosure risk assessed through attacks. Additionally, synthesis introduces randomness into both the sampling starting state and intermediate sampling steps, with the aim of minimizing disclosure risk arising from links between synthetic and reference records.

Synthetic visits generated with limited evidence could supplement clinical AI evaluation, including assessments across clinically defined subgroups, when access to source records is restricted. Our experiments evaluated classifiers trained on synthetic data and tested on real data; the use of synthetic visits as evaluation inputs remains to be validated. Further work should test whether evaluations on synthetic and real data yield consistent conclusions and assess disclosure risk in the intended setting of use.

This study has several limitations. Data were drawn from MIMIC-IV and Duke Health, and applicability to other institutions and populations requires further validation. Mapping tables used during preprocessing grouped clinical codes into coarser categories, which may limit analyses requiring finer clinical detail. In seeking to preserve source-data statistics, the framework may also carry forward data-quality issues such as randomly missing entries and residual outliers. The model generates single visits without capturing temporal dependencies across visits, limiting its use for longitudinal disease and treatment trajectories. Individual-level conditioning information was not matched across synthesis methods. The observed performance differences therefore reflect the complete evidence-conditioned framework and do not isolate the contribution of the diffusion bridge. Results are reported as point estimates without confidence intervals. Privacy was assessed empirically using membership and attribute inference attacks, and EHRDiB provides no formal privacy guarantee.

Future work could integrate differential privacy into evidence-conditioned generation to assess utility and disclosure risk under formal privacy constraints. Extending the framework to multiple visits could enable modeling of longitudinal patient trajectories. Federated learning could support cross-hospital training without centrally pooling patient records, while requiring appropriate privacy protection for model updates.

\label{sec:conclusion}

EHRDiB unifies EHR visit completion and synthesis through a shared representation and an evidence-conditioned latent diffusion bridge, with task-specific rules for retaining supplied information. Across MIMIC-IV and Duke Health, the framework supported downstream tasks while maintaining distributional fidelity and diversity; the evidence-fraction analysis further showed that more supplied evidence was associated with a stronger membership-inference signal. Explicit control over the amount of supplied evidence provides a basis for matching generated EHR data to its intended setting of use.

%% file: sections/methods.tex
\label{sec:methods}

\subsection{Data Sources}
\label{sec:data}

This study used visit-level EHR data from MIMIC-IV and Duke Health. MIMIC-IV version 3.1 is a publicly available, deidentified database covering care at Beth Israel Deaconess Medical Center from 2008 to 2022 \cite{mimiciv,mimiciv31}. We included diagnosis, procedure, medication, and clinical measurement data in this study. The Duke Health data comprised diagnosis, medication, and clinical measurement records from 2015 to 2018 for patients in the Duke PREVENT cohort, which selected individuals aged 30-79 years without established cardiovascular disease at the study baseline \cite{dukeprevent}. In the numerical Duke Health representation (DH), clinical measurement values and their presence indicators were encoded separately. In the discretized representation (DH-D), measurement values were binned into categories, and missing measurements were represented by an explicit missing state. Data from each source were partitioned by patient into training, validation, and test sets in an 80:10:10 ratio.

\subsection{Visit Representation and Task Definition}
\label{sec:formulation}

To define completion and synthesis on these data, a visit is represented by \(K\) modeling groups, denoted by \(\mathbf x=(\mathbf x^{(1)},\ldots,\mathbf x^{(K)})\). Each modeling group corresponds to a distinct combination of clinical modality and variable type present in the dataset. For example, categorical and numerical variables within the same clinical modality form separate modeling groups. Categorical variables have finite state spaces that include an explicit absence state. A numerical group represents the binary presence state and value of each measurement separately. A binary availability mask \(\mathbf m\) indicates which entries of the visit representation are observed. Measurement presence, in contrast, indicates whether a measurement is recorded. For example, when the absence of a laboratory result is known, its presence indicator is observed. The visit representation \(\mathbf x\) and its availability mask \(\mathbf m\) together form the data view.

Generation is conditioned on a subset of the available information, termed an evidence view:
\begin{equation}
\mathbf x_e=\mathbf m_e\odot\mathbf x,\qquad \mathbf m_e\preceq\mathbf m,
\end{equation}
where \(\mathbf m_e\) identifies the supplied variables, \(\odot\) denotes elementwise multiplication, and \(\mathbf m_e\preceq\mathbf m\) restricts selection to available information. The variables preserve the original record state, while the mask marks which information is supplied to the generator. To construct this mask, eligible clinical variables are randomly selected within each group at a specified evidence fraction, which controls the amount of input information. In addition to the randomly selected variables, we always include a diagnosis code recorded for the same visit to provide reliable clinical context. For MIMIC-IV, we use the recorded principal diagnosis. For the Duke Health data, we assign a proxy principal diagnosis by selecting, among the eligible diagnoses recorded for that visit, the one occurring most frequently across training visits. Variable encoding and evidence-selection rules are detailed in Supplementary Methods~\ref{supp:input}.

For both tasks, we denote EHRDiB's conditional generative distribution by \(p_\Theta(\mathbf x\mid\mathbf x_e,\mathbf m_e)\), where \(\Theta\) denotes the model parameters learned from training data. This distribution is induced by the encoder-decoder architecture and latent diffusion bridge described below. Completion uses the evidence view of the target visit to generate a candidate record, then restores all supplied information to obtain the completed record:
\begin{equation}
\widehat{\mathbf x}\sim p_\Theta(\mathbf x\mid\mathbf x_e,\mathbf m_e),\qquad
\widetilde{\mathbf x}=\mathbf x_e+(\mathbf 1-\mathbf m_e)\odot\widehat{\mathbf x}.
\label{eq:covering}
\end{equation}
Here, \(\widehat{\mathbf x}\) is the generated record and \(\widetilde{\mathbf x}\) is the completed record. Synthesis uses a reference population \(\mathcal R\) to generate new visits containing a specified principal diagnosis. Unless otherwise specified, reference visits are drawn from the training set. A reference visit \(\mathbf r\in\mathcal R\) with that diagnosis supplies the evidence view used to condition generation:
\begin{equation}
\widehat{\mathbf x}\sim p_\Theta(\mathbf x\mid\mathbf r_e,\mathbf m_e),\qquad
\mathbf r_e=\mathbf m_e\odot\mathbf r.
\end{equation}
Only the principal diagnosis is explicitly retained in the synthetic record; all other variables are generated, including those supplied as conditioning information.

\subsection{Clinically Structured Representation}
\label{sec:representation}

EHRDiB uses the same group-specific encoders for the data view and the evidence view to construct representations in a common latent space. Each encoder maps clinical variables and their availability mask to a learned Gaussian factor over \(\mathbf z\in\mathbb R^{d_z}\), parameterized by a mean vector and a diagonal covariance matrix. Multiplicative gating incorporates clinical information availability into the encoding alongside variable values. Binary and categorical groups each use one encoder, whereas a numerical group uses separate encoders for measurement presence and value. These two branches remain within the same modeling group, and the value branch receives only present, available results. For categorical variables with an explicit absence state, non-absence categories are represented by indicators, while absence is represented by an all-zero vector with a zero encoding mask. The absence state remains a reconstruction target in the data view.

For each data or evidence view, clinical context fusion (CCF) combines the available encoder factors across subsets of modeling groups. Each valid subset includes the diagnosis group containing the principal diagnosis, providing a shared clinical context:
\begin{equation}
\mathcal S_v=\left\{S\in\mathcal P(\mathcal K_v)\;\middle|\;k_{\mathrm{diag}}\in S\right\},
\end{equation}
where \(\mathcal K_v\) indexes groups with at least one available encoder factor, \(\mathcal P(\mathcal K_v)\) denotes its power set, and \(k_{\mathrm{diag}}\) identifies the diagnosis group. Within each subset, a product of experts combines the available Gaussian factors with a single standard normal prior \(p(\mathbf z)\) to form a Gaussian subset posterior. The presence and value factors of a numerical group remain within the same modeling group. An equal-weight mixture of these subset posteriors gives the variational posterior for the view \cite{mvae,mopoe}:
\begin{equation}
q_\phi(\mathbf z\mid\mathbf x_v,\mathbf m_v)
=\frac{1}{|\mathcal S_v|}\sum_{S\in\mathcal S_v}
\frac{p(\mathbf z)}{Z_{S,v}}
\prod_{k\in S}\prod_{r\in\mathcal E_{k,v}}
q_{\phi_{k,r}}\bigl(\mathbf z\mid\mathbf x_v^{(k)},\mathbf m_v^{(k)}\bigr).
\end{equation}
Here, \(v\) identifies the data or evidence view, and \(\phi\) denotes encoder parameters. Within each group, \(r\in\mathcal E_{k,v}\) indexes the available encoders, whose Gaussian factors are denoted by \(q_{\phi_{k,r}}\). The constant \(Z_{S,v}\) normalizes the product of the prior and all included encoder factors.

Parallel decoders map the shared latent representation \(\mathbf z\) to observation distributions for their respective modeling groups. The groups are assumed conditionally independent given \(\mathbf z\), with their dependencies captured through the shared latent representation. Binary variables use Bernoulli distributions, and categorical variables use distributions over all categories, including the explicit absence state \cite{hivae}. For numerical groups, separate decoder branches model measurement presence and value. Measurement presence follows a Bernoulli distribution, while present values follow a Gaussian distribution. Further methodological details, including the variational network architecture, are provided in Supplementary Methods~\ref{supp:representation}.

\subsection{Evidence-Conditioned Latent Bridge}
\label{sec:bridge}

In the shared latent space, paired data and evidence views of the same visit define the data endpoint \(\mathbf z_0\) and the evidence endpoint \(\mathbf z_T\), respectively. Each endpoint is the posterior mean of the subset containing all available modeling groups in the corresponding view. Conditioning a zero-drift reference diffusion on these paired endpoints yields a Gaussian bridge marginal with the following analytic form \cite{i2sb}:
\begin{equation}
q(\mathbf z_t\mid\mathbf z_0,\mathbf z_T)
=\mathcal N\big(a_t\mathbf z_0+b_t\mathbf z_T,\,c_t^2\mathbf I\big),
\qquad 0<t<T,
\end{equation}
where
\begin{equation}
a_t=\frac{\bar\sigma_t^2}{\sigma_t^2+\bar\sigma_t^2},\qquad
b_t=\frac{\sigma_t^2}{\sigma_t^2+\bar\sigma_t^2},\qquad
c_t^2=\frac{\sigma_t^2\bar\sigma_t^2}{\sigma_t^2+\bar\sigma_t^2}.
\end{equation}
Here, \(\sigma_t^2=\int_0^t\beta(\tau)\,d\tau\) and \(\bar\sigma_t^2=\int_t^T\beta(\tau)\,d\tau\) are the diffusion variances accumulated over the two intervals, with \(\beta(t)>0\) specifying the diffusion variance rate. Completion uses the evidence-endpoint distribution directly as its starting distribution. Synthesis applies additive Gaussian perturbation to this distribution, using the same perturbation rule during training and generation \cite{ecsi}. The resulting perturbed state replaces the evidence endpoint as the terminal boundary in the bridge distribution above.

The analytic bridge marginal above assumes that both endpoints are known. During generation, however, the data endpoint is unknown, and reverse sampling requires the conditional score. EHRDiB estimates this score through endpoint prediction \cite{ddbm}. A conditional endpoint network \(F_\theta(\mathbf z_t,t,\mathbf z_T)\), with parameters \(\theta\), estimates \(\mathbf z_0\) from the intermediate latent state, diffusion time, and unperturbed evidence endpoint. Time embeddings encode the position along the bridge, while evidence-conditioned feature-wise modulation incorporates the clinical context into the hidden representations \cite{film}. For the unperturbed bridge, the analytic relationship yields the score estimate:
\begin{equation}
\mathbf s_\theta
=\tfrac{a_t}{c_t^2}F_\theta(\mathbf z_t,t,\mathbf z_T)
+\tfrac{b_t}{c_t^2}\mathbf z_T
-\tfrac{1}{c_t^2}\mathbf z_t.
\label{eq:bridge-score-estimate}
\end{equation}
Here, \(\mathbf z_T\) denotes the unperturbed evidence endpoint. For synthesis, the terminal endpoint sampled from the perturbed starting distribution replaces \(\mathbf z_T\) in the second term, while the network remains conditioned on the unperturbed evidence endpoint. Together with the analytic transition score of the reference diffusion, this estimate guides reverse sampling toward time zero. The hybrid sampler alternates stochastic updates with deterministic Heun integration. Starting-distribution perturbation and sampling stochasticity control variation in generation; their effects on diversity and disclosure risk are assessed empirically. Further methodological details, including the endpoint network architecture, are provided in Supplementary Methods~\ref{supp:bridge}.

\subsection{Training and Generation}
\label{sec:training}

Learning this generator involves two stages. First, the encoders and decoders are jointly trained on data views and randomly constructed evidence views. The objective combines type-specific reconstruction losses with Kullback-Leibler regularization toward the standard normal prior. Reconstruction scores only available targets in the current view, and regularization averages over valid subset posteriors rather than taking the exact KL divergence of their mixture. The encoders and decoders are then frozen, and a separate bridge is fitted for each task configuration by weighted regression to the encoded data endpoints.

After reverse integration, the frozen decoders produce the parameters of the observation distributions used to generate a visit. Synthesis is evaluated with two output rules applied to the same trained model: Bernoulli sampling draws binary outcomes and categorical states, whereas calibrated thresholding (Cali-T) uses training-set prevalence to set output thresholds. Numerical values are bounded decoder estimates and are used only when a result is marked present. The task-specific preservation rules are then applied to the generated visit. Further training and generation details are provided in Supplementary Methods~\ref{supp:training}, and the training and sampling settings for the main experiments are summarized in \textbf{Supplementary Tab.~\ref{tab:reference-settings}}.

\subsection{Comparison Methods}
\label{sec:comparators}

To assess EHRDiB against existing approaches, methods operating on the same fixed-dimensional visit representations were evaluated using shared data partitions and measures. Completion was compared with CISS-VAE, GAIN, and CSDI \cite{cissvae,gain,csdi}, with supplied evidence restored in all completed outputs. Synthesis was compared with the variational models MVAE and MoPoE-VAE \cite{mvae,mopoe}, the adversarial models medGAN, medBGAN, and EMR-WGAN \cite{medgan,medwgan,emrwgan}, and the diffusion models TabSyn, TabDiff, and ScoEHR \cite{tabsyn,tabdiff,scoehr}.

\subsection{Evaluation Measures}
\label{sec:evaluation}

Evaluation assessed downstream predictive utility, generative diversity, and distributional fidelity for both completion and synthesis, and disclosure risk for synthesis. Prediction targets, metric formulas, and evaluation settings are detailed in Supplementary Methods~\ref{supp:evaluation}.

Predictive utility assessed whether generated or completed records supported phenotype identification on real patients. Classifiers trained on generated or completed training visits were compared with those trained on real visits, evaluated on the same real test partition \cite{ehrbenchmark}. LightGBM configuration and training-set sizes were matched \cite{lightgbm}. MIMIC-IV used ten principal-diagnosis targets and seven Phecode-based phenotypes; DH and DH-D used ten diagnosis targets. Target-defining diagnosis features were removed from classifier inputs. We report target-averaged differences in AUROC and AUPRC.

Generative diversity was evaluated using the Average Pairwise Distance Ratio (APDR) and Coverage, both computed in a common principal-component space fitted on real training data. APDR compares the mean distance between generated visits with that between real visits; a value near one indicates similar overall dispersion. Coverage measures the proportion of real-data neighborhoods containing at least one generated visit \cite{densitycoverage}. Higher Coverage indicates greater coverage of the real data.

Fidelity assessed the distributions of individual variables and the relationships between them. Feature Distribution Distance (FDD) combines prevalence discrepancies with differences in the distributions of recorded numerical values. Blockwise Correlation Matrix Distance (BCMD) compares correlations within and between clinical modalities. Lower FDD and BCMD indicate smaller discrepancies from real data.

Disclosure risk for synthesis was assessed using a membership inference attack (MIA) and an attribute inference attack (AIA) \cite{shokri2017membership,stadler2022synthetic}. MIA performance is measured by AUROC, which quantifies how well nearest-synthetic-visit distances distinguish patient-disjoint training members from test nonmembers. A value near 0.5 indicates little discrimination under this attack; departures in either direction can carry membership information. For AIA, we compare the AUPRC for inferring hidden diagnosis and medication codes from known patient information, using either synthetic records or an independent, equally sized real auxiliary set as the reference. The reported advantage (\(\Delta\)AIA) is the AUPRC difference between the synthetic and real auxiliary references; positive values indicate improved inference using synthetic data. These attacks do not establish formal privacy guarantees.

\subsection{Evidence-Fraction and Component Analyses}
\label{sec:component-analyses}

Beyond the default task settings, we further varied the evidence fraction in MIMIC-IV synthesis to examine how the amount of supplied patient information affected utility and disclosure risk. Models were trained and evaluated under twelve evidence-fraction intervals. All configurations used the same data partitions, model architecture, training budget, and other experimental settings. Each interval was summarized by its mean evidence fraction and evaluated using the measures described above. The intervals are listed in Supplementary Methods~\ref{supp:evaluation}.

\textcolor{black}{To assess individual design choices, model components were examined in MIMIC-IV synthesis by varying the availability encoding method, the KL weight, the analytical formulation of the bridge path, and the sampler type with the remaining components fixed within each study (Supplementary Methods~\ref{supp:components}). Sampling hyperparameters are evaluated in relation to the specific experimental configuration and the priorities given to fidelity, diversity, and utility.}

%% file: sections/supplement_body.tex
\newcounter{suppmethod}
\renewcommand{\thesuppmethod}{S\arabic{suppmethod}}

\subsection*{Contents}
\begin{figure}[H]
\centering
\includegraphics[width=.97\linewidth]{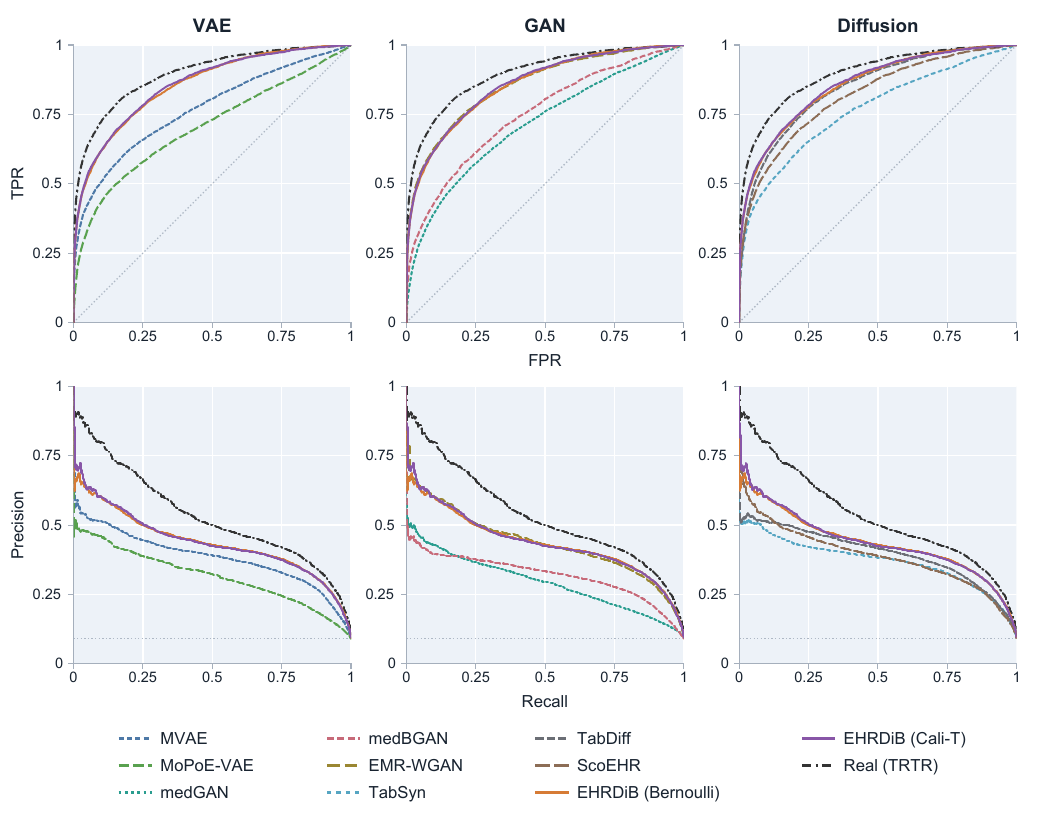}
\caption{MIMIC-IV synthesis utility for the seven Phecode-defined phenotypes. Macro-averaged ROC (top) and precision-recall (bottom) curves compare LightGBM classifiers trained on real (TRTR) or synthetic records and evaluated on the same real test set. Columns group variational, adversarial, and diffusion comparators, with EHRDiB (Bernoulli) and EHRDiB (Cali-T) shown in each.}
\label{fig:utility-b}
\end{figure}

\begin{figure}[H]
\centering
\includegraphics[width=\linewidth]{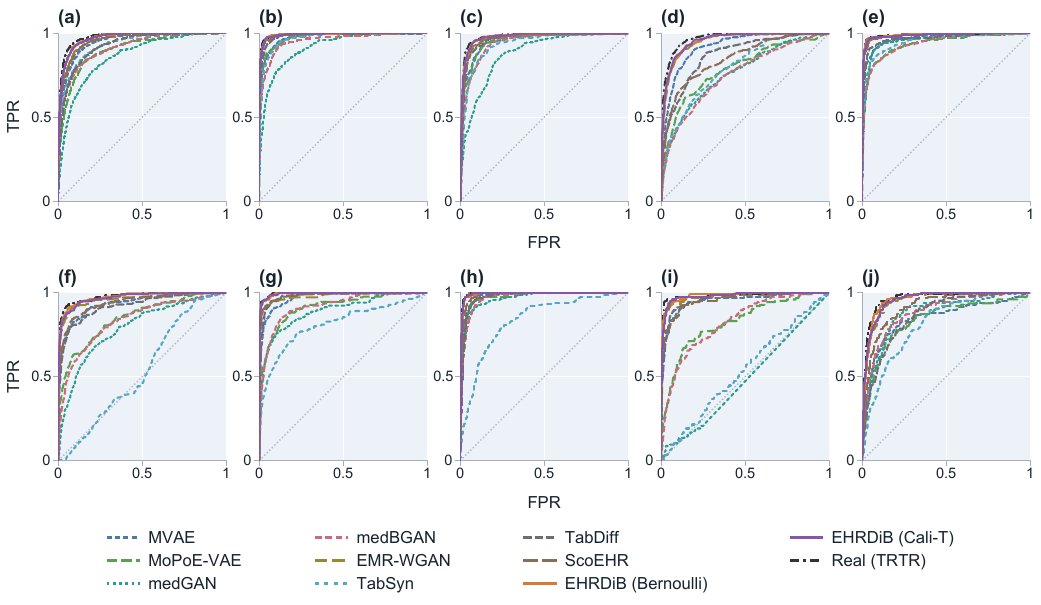}
\caption{MIMIC-IV synthesis ROC curves for the ten principal-diagnosis targets: (a) sepsis; (b) acute myocardial infarction; (c) major depressive disorder; (d) type 2 diabetes mellitus; (e) cerebral infarction; (f) cholelithiasis and cholecystitis; (g) fracture of femur; (h) alcoholic liver disease; (i) malignant neoplasm of bronchus and lung; (j) respiratory failure. Each panel compares LightGBM classifiers trained on real (TRTR) or synthetic records and evaluated on the same real test set. EHRDiB (Bernoulli) and EHRDiB (Cali-T) are shown alongside the comparators. FPR, false positive rate; TPR, true positive rate.}
\label{fig:diagnosis-a}
\end{figure}

\begin{figure}[H]
\centering
\includegraphics[width=\linewidth]{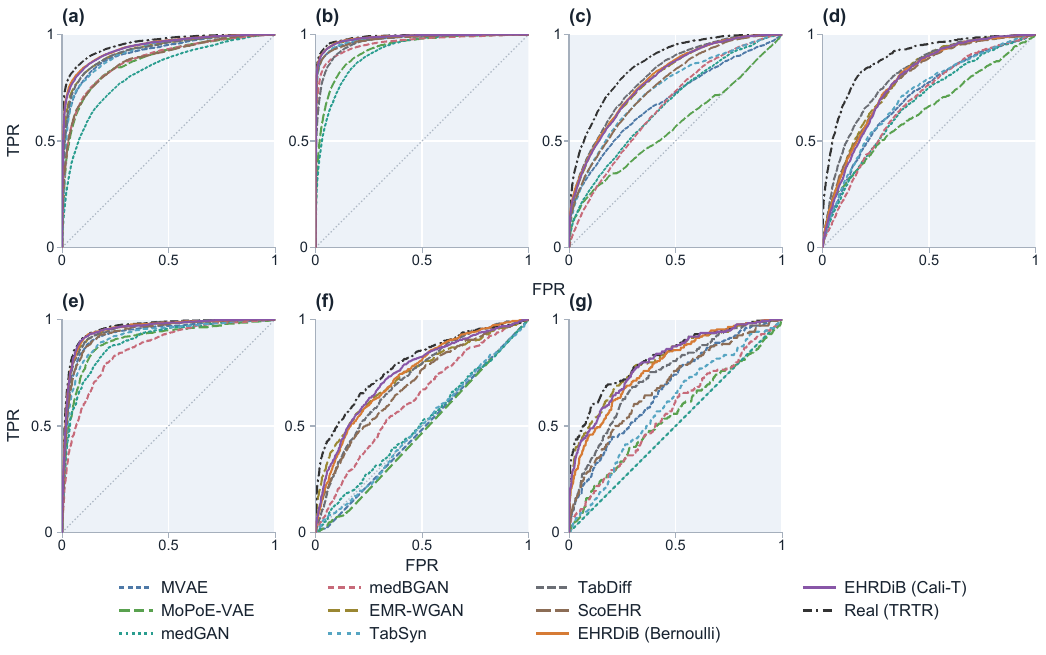}
\caption{MIMIC-IV synthesis ROC curves for the seven Phecode-defined phenotypes: (a) type 2 diabetes mellitus; (b) heart failure; (c) cerebrovascular disease; (d) bipolar disorder; (e) suicidal ideation; (f) rheumatoid arthritis; (g) multiple sclerosis. Each panel compares LightGBM classifiers trained on real (TRTR) or synthetic records and evaluated on the same real test set. EHRDiB (Bernoulli) and EHRDiB (Cali-T) are shown alongside the comparators. FPR, false positive rate; TPR, true positive rate.}
\label{fig:diagnosis-b}
\end{figure}

\begingroup\small
\setlength{\parindent}{0pt}
\textbf{Supplementary Methods}\par
\hyperref[supp:input]{\ref*{supp:input}. Input Representation and Evidence Selection} \dotfill \pageref{supp:input}\par
\hyperref[supp:representation]{\ref*{supp:representation}. Clinically Structured Representation} \dotfill \pageref{supp:representation}\par
\hyperref[supp:bridge]{\ref*{supp:bridge}. Evidence-Conditioned Latent Bridge} \dotfill \pageref{supp:bridge}\par
\hyperref[supp:training]{\ref*{supp:training}. Training and Generation} \dotfill \pageref{supp:training}\par
\hyperref[supp:evaluation]{\ref*{supp:evaluation}. Evaluation Measures and Settings} \dotfill \pageref{supp:evaluation}\par
\hyperref[supp:components]{\ref*{supp:components}. Component Analysis Design} \dotfill \pageref{supp:components}\par
\medskip
\textbf{Supplementary Results}\par
\hyperref[supp:results-population]{Study populations and predictive utility} \dotfill \pageref{supp:results-population}\par
\hyperref[supp:results-evidence]{Evidence-fraction configurations} \dotfill \pageref{supp:results-evidence}\par
\hyperref[supp:results-components]{Component analyses} \dotfill \pageref{supp:results-components}\par
\medskip
\hyperref[supp:tables]{\textbf{Supplementary Tabs.~\ref*{tab:population}-\ref*{tab:reference-settings}}} \dotfill \pageref{supp:tables}\par
\hyperref[supp:figures]{\textbf{Supplementary Figs.~\ref*{fig:utility-b}-\ref*{fig:diagnosis-b}}} \dotfill \pageref{supp:figures}\par
\endgroup
\clearpage
\section*{Supplementary Methods}

\refstepcounter{suppmethod}
\subsection*{Supplementary Methods \thesuppmethod. Input Representation and Evidence Selection}
\label{supp:input}
We analyzed the same Duke Health visits in two representations: Duke Health (DH), which retained clinical measurements as numerical values, and Duke Health Discrete (DH-D), which discretized these values into categories using binning.

For MIMIC-IV, we used the recorded principal diagnosis. For Duke Health, we assigned a proxy principal diagnosis by selecting the eligible visit diagnosis with the highest training-set frequency.

Binary variables use one encoded coordinate each. A categorical variable uses one indicator for each non-absence category, with an all-zero input vector for absence; its decoder still assigns probability to every state. All indicators of a categorical variable share one evidence-selection decision. Numerical presence and value are paired during selection, and a value is eligible only when it is available and its measurement is present. Thus the general mask relation is
\begin{equation}
\mathbf m_{e,\mathrm{val}}^{(k)}
=\mathbf m_{e,\mathrm{pres}}^{(k)}\odot\mathbf x_{\mathrm{pres}}^{(k)}
\odot\mathbf m_{\mathrm{val}}^{(k)}.
\end{equation}
When every recorded present measurement has an available value, the last factor is redundant. An absent categorical variable is supervised as an output state in a data view but is excluded from randomly selected categorical evidence. The encoder-eligibility rules in Supplementary Methods~\ref{supp:representation} implement this distinction.

Evidence selection draws one ratio per visit and applies it to every eligible group. With \(n>1\) eligible variables, the selected count is the rounded ratio times \(n\), clipped to \([1,n-1]\); random ranking chooses the variables. A single eligible variable is retained by a Bernoulli draw with that ratio. Empty groups supply no evidence. Categorical absence states are ineligible, numerical presence and value are paired, and the principal diagnosis is added after selection.

\refstepcounter{suppmethod}
\subsection*{Supplementary Methods \thesuppmethod. Clinically Structured Representation}
\label{supp:representation}
\paragraph{Architecture Overview}
EHRDiB has two sequentially trained neural systems. The variational network encodes data and evidence views into a shared \(d_z\)-dimensional space and parameterizes typed output likelihoods. It is then frozen while the endpoint network learns the evidence-to-data bridge.

\paragraph{Variational Representation Network}
Each modeling group has its own encoder parameters. Let \(\widetilde{\mathbf m}^{(k)}_v\) be its encoder-eligibility mask. For binary and numerical branches it is the aligned view mask; for a categorical variable with absence state zero, its logical-variable entry is
\begin{equation}
\widetilde m_{v,j}^{(k)}=m_{v,j}^{(k)}\mathbb I[x_{v,j}^{(k)}\ne0],
\end{equation}
expanded across its non-absence input indicators. This suppresses absent categorical variables at the encoder while leaving their data-view reconstruction targets active. Numerical value eligibility additionally requires an available, positive measurement-presence indicator.

With \(\overline{\mathbf x}^{(k)}_v=\mathbf x^{(k)}_v\odot\widetilde{\mathbf m}^{(k)}_v\), the first hidden layer is
\begin{equation}
\begin{aligned}
\mathbf a^{(k)}_1&=\operatorname{SiLU}(W_x^{(k)}\overline{\mathbf x}^{(k)}_v+\mathbf b_x^{(k)}),\\
\mathbf g^{(k)}&=\mathbf1+\tanh(W_m^{(k)}\widetilde{\mathbf m}^{(k)}_v+\mathbf b_m^{(k)}),\\
\mathbf h^{(k)}_1&=\operatorname{Dropout}(\mathbf a^{(k)}_1\odot\mathbf g^{(k)}).
\end{aligned}
\end{equation}
Both projections map encoded coordinates to the first hidden width. Zero-initialized gate weights and biases give \(\mathbf g^{(k)}=\mathbf1\) initially. Linear-SiLU-dropout layers then feed independent posterior-mean and clipped-log-variance heads. The same network processes data and evidence views.

Numerical presence and value have separate encoders and remain inner experts of one numerical modeling group. Categorical and numerical measurements form different outer groups when both are present. Analytic fusion includes a standard-normal prior in every valid diagnosis-containing subset product, then mixes the subset posteriors with equal weights \cite{mvae,mopoe}. Balanced component selection is used for variational learning; the largest valid subset mean defines each data or evidence endpoint.

Every decoder branch receives the same latent vector and has a separate linear-SiLU-dropout trunk. Its head produces binary logits, grouped categorical logits, numerical-presence logits, or numerical values. The output rules and task-level preservation are specified in Supplementary Methods~\ref{supp:training}.

\paragraph{Gaussian Fusion}

Let \(\mathcal E_{k,v}\) denote the available encoder experts within modeling group \(k\) for view \(v\). For a valid subset \(S\), \(q_{\phi,S}(\mathbf z\mid\mathbf x_{v,S},\mathbf m_{v,S})\) denotes the normalized product of its available encoder factors and the standard normal prior. This subset posterior has diagonal Gaussian parameters
\begin{equation}
\begin{split}
\boldsymbol\lambda_{S,v}&=\mathbf1+
\sum_{k\in S}\sum_{r\in\mathcal E_{k,v}}\boldsymbol\sigma_{k,r,v}^{-2},
\qquad \boldsymbol\sigma_{S,v}^2=\boldsymbol\lambda_{S,v}^{-1},\\
\boldsymbol\mu_{S,v}&=\boldsymbol\sigma_{S,v}^2\odot
\sum_{k\in S}\sum_{r\in\mathcal E_{k,v}}
\boldsymbol\mu_{k,r,v}\odot\boldsymbol\sigma_{k,r,v}^{-2}.
\end{split}
\end{equation}
The encoder factors are learned jointly. A single standard-normal prior contributes the unit precision term.

\paragraph{Numerical Observation Likelihood}
In numerical group \(k\), \(\mathbf x^{(k)}_{\mathrm{pres}}\in\{0,1\}^{D_k}\) records binary presence for \(D_k\) measurements, and \(\mathbf x^{(k)}_{\mathrm{val}}\in\mathbb R^{D_k}\) contains their values. For the preprocessed data views, in which presence indicators are observed and present measurements have available values, the likelihood factorizes as
\begin{equation}
\begin{aligned}
&p_{\psi_k}\bigl(\mathbf x^{(k)}_{\mathrm{pres}},\mathbf x^{(k)}_{\mathrm{val,obs}}\mid\mathbf z\bigr)\\
&\quad=p_{\psi_{k,\mathrm{pres}}}\bigl(\mathbf x^{(k)}_{\mathrm{pres}}\mid\mathbf z\bigr)
\prod_{j:\,x^{(k)}_{\mathrm{pres},j}=1}
\mathcal N\bigl(x^{(k)}_{\mathrm{val},j};\,\widehat x^{(k)}_{\mathrm{val},j}(\mathbf z),\,\sigma^2_{\mathrm{val}}\bigr).
\end{aligned}
\label{eq:numerical-likelihood}
\end{equation}
Here, \(\psi_k\) denotes the decoder parameters, and \(\mathbf x^{(k)}_{\mathrm{val,obs}}\) contains the values of present measurements. The presence likelihood is Bernoulli, while the value likelihood is Gaussian with decoder-predicted mean \(\widehat x^{(k)}_{\mathrm{val},j}(\mathbf z)\) and fixed variance \(\sigma^2_{\mathrm{val}}\). Thus, a known absence contributes to the presence likelihood without being assigned a numerical-value likelihood.

\refstepcounter{suppmethod}
\subsection*{Supplementary Methods \thesuppmethod. Evidence-Conditioned Latent Bridge}
\label{supp:bridge}
\paragraph{Forward Bridge Process}
Let \(E_\phi\) denote expert encoding, clinical context fusion, and extraction of the posterior mean of \(q_{\phi,\mathcal K_v}\) for the full available subset \(\mathcal K_v\). Two views of the same visit define paired endpoints:
\begin{equation}
\mathbf z_0=E_\phi(\mathbf x,\mathbf m),\qquad
\mathbf z_T=E_\phi(\mathbf x_e,\mathbf m_e).
\end{equation}
The data endpoint \(\mathbf z_0\) represents the target visit, and the evidence endpoint \(\mathbf z_T\) conditions generation. The starting state \(\mathbf z_T^{\mathrm{start}}\) equals the evidence endpoint for completion and a perturbed version for synthesis. The symbol \(\mathbf z_T\) always denotes the unperturbed evidence representation; the path boundary at time \(T\) is \(\mathbf z_T^{\mathrm{start}}\). Bridge time parameterizes the latent diffusion trajectory. Reverse sampling proceeds from the terminal boundary toward time zero.

The reference process is a zero-drift diffusion,
\begin{equation}
d\mathbf z_t=\sqrt{\beta(t)}\,d\mathbf W_t,\qquad
\sigma_t^2=\int_0^t\beta(\tau)\,d\tau,\qquad
\bar\sigma_t^2=\int_t^T\beta(\tau)\,d\tau,
\end{equation}
where \(\mathbf W_t\) is a standard \(d_z\)-dimensional Wiener process and \(\beta(t)>0\) is an integrable diffusion variance rate on \([0,T]\). Conditioning this reference on the endpoints gives a Gaussian bridge marginal for \(0<t<T\) \cite{i2sb}:
\begin{equation}
q(\mathbf z_t\mid\mathbf z_0,\mathbf z_T^{\mathrm{start}})
=\mathcal N(a_t\mathbf z_0+b_t\mathbf z_T^{\mathrm{start}},c_t^2\mathbf I),
\end{equation}
\begin{equation}
a_t=\frac{\bar\sigma_t^2}{\sigma_t^2+\bar\sigma_t^2},\qquad
b_t=\frac{\sigma_t^2}{\sigma_t^2+\bar\sigma_t^2},\qquad
c_t^2=\frac{\sigma_t^2\bar\sigma_t^2}{\sigma_t^2+\bar\sigma_t^2}.
\end{equation}
For this continuous reference, \(a_t+b_t=1\) and \(c_t^2>0\) in the interior; the distributions converge to point masses at the prescribed endpoints as \(t\to0\) and \(t\to T\).

\paragraph{Endpoint Prediction and Conditional Score}
\label{app:conditional-score}
The endpoint predictor estimates the data representation from an intermediate state, time, and the unperturbed evidence endpoint:
\begin{equation}
\widehat{\mathbf z}_0=F_\theta(\mathbf z_t,t,\mathbf z_T).
\end{equation}
The predictor is trained by weighted endpoint regression (Supplementary Methods~\ref{supp:training}).

For the continuous reference, let \(q_t(\mathbf z_t\mid\mathbf z_T^{\mathrm{start}},\mathbf z_T)\) denote the bridge marginal after integrating over data endpoints. Differentiating this Gaussian mixture gives the exact conditional score; replacing its conditional mean by the endpoint prediction gives the score used by the model \cite{ddbm}:
\begin{equation}
\begin{split}
\mathbf s^*&=\nabla_{\mathbf z_t}\log q_t(\mathbf z_t\mid\mathbf z_T^{\mathrm{start}},\mathbf z_T)\\
&=-\frac{\mathbf z_t-a_t\mathbb E[\mathbf z_0\mid\mathbf z_t,t,\mathbf z_T^{\mathrm{start}},\mathbf z_T]
-b_t\mathbf z_T^{\mathrm{start}}}{c_t^2},\\
\mathbf s_\theta(\mathbf z_t,t,\mathbf z_T^{\mathrm{start}},\mathbf z_T)
&=-\frac{\mathbf z_t-a_t\widehat{\mathbf z}_0-b_t\mathbf z_T^{\mathrm{start}}}{c_t^2}.
\end{split}
\label{eq:plugin-score}
\end{equation}
When the starting state equals the evidence endpoint, endpoint regression under the corresponding bridge marginal targets the conditional mean in the score identity. For synthesis, \(F_\theta\) conditions on the intermediate state, time, and unperturbed evidence endpoint. Substituting this evidence-conditioned prediction into the terminal-conditioned score formula defines the plug-in approximation used for sampling.

For independent Gaussian endpoint perturbation, marginalizing the perturbed boundary gives
\begin{equation}
\begin{split}
q(\mathbf z_t\mid\mathbf z_0,\mathbf z_T)
&=\mathcal N\!\left(a_t\mathbf z_0+b_t\mathbf z_T,
 (c_t^2+b_t^2\sigma_{\mathrm{start}}^2)\mathbf I\right),\\
\nabla_{\mathbf z_t}\log q_t(\mathbf z_t\mid\mathbf z_T)
&=-\frac{\mathbf z_t-a_t\mathbb E[\mathbf z_0\mid\mathbf z_t,t,\mathbf z_T]-b_t\mathbf z_T}
{c_t^2+b_t^2\sigma_{\mathrm{start}}^2}.
\end{split}
\end{equation}
The added variance follows from Gaussian convolution. Integrating over data endpoints gives the evidence-conditioned score above, using the conditional-score argument of DDBM and ECSI \cite{ddbm,ecsi}. At zero endpoint perturbation, the terminal-conditioned and perturbation-marginalized score identities coincide. Sampling uses Eq.~\eqref{eq:plugin-score} with the finite coefficients defined below.

\paragraph{Conditional Endpoint Prediction Network}
The analytic wrapper takes \(\mathbf z_t\), \(t>0\), and \(\mathbf z_T\), supplies the coefficients defined under \hyperref[app:preconditioning]{Endpoint Preconditioning} below, and combines the residual network output with the state skip term. The trainable network concatenates \(c_{\mathrm{in}}(t)\mathbf z_t\) and \(\mathbf z_T\) once at its input, giving a \(2d_z\)-dimensional vector projected to the first hidden width.

A fixed sinusoidal encoding of \(c_{\mathrm{noise}}(t)\), followed by a two-layer SiLU MLP, gives a time embedding \(\mathbf e_t\) \cite{ddpm,ddbm}. A separate two-layer SiLU MLP maps \(\mathbf z_T\) to the evidence embedding \(\mathbf e_e\). Both embeddings have dimension \(d_t\) and are reused in every residual block. The evidence embedding provides affine scale and shift, adapting feature-wise conditioning \cite{film}.

For block \(l\), the exact operation order is
\begin{equation}
\begin{aligned}
\mathbf u_l&=W_l^{\mathrm{in}}\operatorname{SiLU}(\operatorname{LN}(\mathbf h_l))
+\mathbf b_l^{\mathrm{in}}+A_l\operatorname{SiLU}(\mathbf e_t)+\mathbf b_l^t,\\
[\boldsymbol\gamma_l\mathbin{\|}\boldsymbol\delta_l]
&=B_l\operatorname{SiLU}(\mathbf e_e)+\mathbf b_l^e,\\
\widetilde{\mathbf u}_l&=(\mathbf1+\boldsymbol\gamma_l)\odot\operatorname{LN}(\mathbf u_l)+\boldsymbol\delta_l,\\
\mathbf h_{l+1}&=P_l\mathbf h_l+
W_l^{\mathrm{out}}\operatorname{Dropout}(\operatorname{SiLU}(\widetilde{\mathbf u}_l))
+\mathbf b_l^{\mathrm{out}}.
\end{aligned}
\end{equation}
The shortcut \(P_l\) is identity at equal width and a learned bias-free projection otherwise. The residual output weights and bias start at zero. Hidden widths contract to a central bottleneck and expand symmetrically. A final LayerNorm-SiLU-linear head, also zero-initialized, returns the \(d_z\)-dimensional residual prediction.

\paragraph{Reverse Bridge Dynamics}
For \(0<t<T\), the continuous reference has transition score toward the starting state
\begin{equation}
\mathbf s_T(\mathbf z_t,t,\mathbf z_T^{\mathrm{start}})
=\nabla_{\mathbf z_t}\log p_{\mathrm{ref}}(\mathbf z_T^{\mathrm{start}}\mid\mathbf z_t)
=-\frac{\mathbf z_t-\mathbf z_T^{\mathrm{start}}}{\bar\sigma_t^2}.
\end{equation}
Here \(p_{\mathrm{ref}}\) is the reference diffusion's transition density from \(t\) to \(T\). With the exact bridge score, time reversal gives the stochastic and probability-flow dynamics \cite{ddbm}. Substituting \(\mathbf s_\theta\) gives their estimated drift forms:
\begin{equation}
\mathbf u_\theta^{\mathrm{SDE}}=\beta(t)(\mathbf s_T-\mathbf s_\theta),\qquad
\mathbf v_\theta=\beta(t)(\mathbf s_T-\tfrac12\mathbf s_\theta),\qquad
g^2(t)=\beta(t).
\label{eq:reverse-dynamics}
\end{equation}
Here \(g(t)\) is the stochastic diffusion amplitude. These drifts use decreasing time, \(dt<0\), while stochastic increments have covariance \(g^2(t)|dt|\mathbf I\). Reverse sampling combines stochastic updates with deterministic integration toward time zero. The finite approximation and numerical solver are specified below.

\paragraph{Finite Bridge Approximation}
\label{app:finite-bridge}

The implementation uses a finite symmetric variance schedule. For positive increments \(\delta\beta_0,\ldots,\delta\beta_{N-1}\), define
\begin{equation}
U_n=\sum_{j=0}^{n}\delta\beta_j,\qquad
V_n=\sum_{j=n}^{N-1}\delta\beta_j,\qquad
V=\sum_{j=0}^{N-1}\delta\beta_j.
\end{equation}
With normalized time \(T=1\) and \(n(t)=\operatorname{round}((N-1)t)\), the implemented coefficients are
\begin{equation}
a_t^{\mathrm{disc}}=\frac{V_{n(t)}}{V},\qquad b_t^{\mathrm{disc}}=\frac{U_{n(t)}}{V},\qquad
(c_t^{\mathrm{disc}})^2=\frac{U_{n(t)}V_{n(t)}}{V},\qquad
\beta_{\mathrm{disc}}(t)=(N-1)\delta\beta_{n(t)}.
\label{eq:finite-bridge}
\end{equation}
Both cumulative sums include the current increment, giving \(U_n+V_n=V+\delta\beta_n\). The superscript distinguishes the finite coefficients used during training and sampling from the continuous reference in the main text. Intermediate training states are sampled directly:
\begin{equation}
\begin{split}
q^{\mathrm{disc}}(\mathbf z_t\mid\mathbf z_0,\mathbf z_T^{\mathrm{start}})
&=\mathcal N(a_t^{\mathrm{disc}}\mathbf z_0+b_t^{\mathrm{disc}}\mathbf z_T^{\mathrm{start}},(c_t^{\mathrm{disc}})^2\mathbf I),\\
\mathbf z_t&=a_t^{\mathrm{disc}}\mathbf z_0+b_t^{\mathrm{disc}}\mathbf z_T^{\mathrm{start}}+c_t^{\mathrm{disc}}\boldsymbol\epsilon,
\quad\boldsymbol\epsilon\sim\mathcal N(\mathbf0,\mathbf I).
\end{split}
\end{equation}

The conditional endpoint network is fitted to these finite-schedule states. In the score used for sampling, \(a_t,b_t,c_t\) in Eq.~\eqref{eq:plugin-score} are replaced by \(a_t^{\mathrm{disc}},b_t^{\mathrm{disc}},c_t^{\mathrm{disc}}\). The shared increment in the cumulative sums makes \(a_t^{\mathrm{disc}}+b_t^{\mathrm{disc}}>1\) and leaves nonzero conditional variance at the finite schedule boundaries.

\paragraph{Endpoint Preconditioning}
\label{app:preconditioning}

In the preconditioning expressions below, \(a_t,b_t,c_t\) abbreviate the finite coefficients in Eq.~\eqref{eq:finite-bridge}.
Analytic preconditioning separates the residual network from endpoint prediction \cite{ddbm,edm}. Let \(\nu_0,\nu_T>0\) and \(\chi\) be fixed scale and cross-covariance hyperparameters, with \(|\chi|<\nu_0\nu_T\). For preconditioning, we set the effective evidence scale to \(\nu_T^{\mathrm{eff}}=\nu_T+\sigma_{\mathrm{start}}\), where \(\sigma_{\mathrm{start}}\geq0\) is the endpoint perturbation scale. Define
\begin{equation}
\begin{split}
v_t&=a_t^2\nu_0^2+b_t^2(\nu_T^{\mathrm{eff}})^2+2a_tb_t\chi+c_t^2,\\
r_t&=b_t^2\!\left(\nu_0^2(\nu_T^{\mathrm{eff}})^2-\chi^2\right)+\nu_0^2c_t^2,\\
c_{\mathrm{in}}(t)&=v_t^{-1/2},\qquad
c_{\mathrm{skip}}(t)=\frac{a_t\nu_0^2+b_t\chi}{v_t},\\
c_{\mathrm{out}}(t)&=\sqrt{r_t/v_t},\qquad
c_{\mathrm{noise}}(t)=\kappa\log t,\quad \kappa>0,\ t>0.
\end{split}
\end{equation}
The positive residual scale defines the predictor:
\begin{equation}
F_\theta(\mathbf z_t,t,\mathbf z_T)
=c_{\mathrm{skip}}(t)\mathbf z_t+c_{\mathrm{out}}(t)
f_\theta(c_{\mathrm{in}}(t)\mathbf z_t,c_{\mathrm{noise}}(t),\mathbf z_T),
\end{equation}

\paragraph{Numerical Reverse Integration}
\label{app:reverse-integration}

Let \(\mathbf s_\theta^{\mathrm{disc}}\) be Eq.~\eqref{eq:plugin-score} evaluated with the finite bridge coefficients. The transition score and drifts used by the sampler are
\begin{equation}
\begin{split}
\mathbf s_T^{\mathrm{disc}}&=-\frac{\mathbf z_t-\mathbf z_T^{\mathrm{start}}}{V_{n(t)}},\\
\mathbf u_\theta^{\mathrm{disc}}&=\beta_{\mathrm{disc}}(t)
(\mathbf s_T^{\mathrm{disc}}-\mathbf s_\theta^{\mathrm{disc}}),\\
\mathbf v_\theta^{\mathrm{disc}}&=\beta_{\mathrm{disc}}(t)
(\mathbf s_T^{\mathrm{disc}}-\tfrac12\mathbf s_\theta^{\mathrm{disc}}),
\qquad g_{\mathrm{disc}}^2(t)=\beta_{\mathrm{disc}}(t).
\end{split}
\end{equation}
These drifts evaluate the reference drift forms in Eq.~\eqref{eq:reverse-dynamics} with the finite bridge coefficients at the current state and time.

On a decreasing grid \(t_i>t_{i+1}\), a stochasticity parameter \(\omega_i\in[0,1)\) allocates part of an interval to an Euler-Maruyama substep:
\begin{equation}
\begin{split}
\widehat t_i&=t_i+\omega_i(t_{i+1}-t_i),\qquad
\delta_i^{\mathrm S}=\widehat t_i-t_i\leq0,\\
\widehat{\mathbf z}_i&=\mathbf z_{t_i}+
\delta_i^{\mathrm S}\mathbf u_\theta^{\mathrm{disc}}(\mathbf z_{t_i},t_i,\mathbf z_T^{\mathrm{start}},\mathbf z_T)
+\sqrt{|\delta_i^{\mathrm S}|}\,g_{\mathrm{disc}}(t_i)\boldsymbol\xi_i,
\quad\boldsymbol\xi_i\sim\mathcal N(\mathbf0,\mathbf I).
\end{split}
\end{equation}
At zero stochasticity this leaves both time and state unchanged. A Heun step integrates the remaining interval, \(\Delta t_i=t_{i+1}-\widehat t_i\):
\begin{equation}
\begin{split}
\widetilde{\mathbf z}_{i+1}&=\widehat{\mathbf z}_i+
\Delta t_i\mathbf v_\theta^{\mathrm{disc}}(\widehat{\mathbf z}_i,\widehat t_i,\mathbf z_T^{\mathrm{start}},\mathbf z_T),\\
\mathbf z_{t_{i+1}}&=\widehat{\mathbf z}_i+\frac{\Delta t_i}{2}
\left[\mathbf v_\theta^{\mathrm{disc}}(\widehat{\mathbf z}_i,\widehat t_i,\mathbf z_T^{\mathrm{start}},\mathbf z_T)
+\mathbf v_\theta^{\mathrm{disc}}(\widetilde{\mathbf z}_{i+1},t_{i+1},\mathbf z_T^{\mathrm{start}},\mathbf z_T)\right].
\end{split}
\end{equation}
At the final interval ending at zero, only the Euler predictor is used, avoiding a network evaluation at zero time. Sampling starts below \(T\) with the starting state as its initial state and proceeds toward the final integrated state \(\mathbf z_0^{\mathrm{gen}}\). The sampler alternates stochastic updates and deterministic integration \cite{ddbm}, using a second-order Heun update for the latter \cite{edm}. Stochasticity controls path randomness; the solver controls deterministic integration error.

The final integrated latent state \(\mathbf z_0^{\mathrm{gen}}\) is passed to the frozen decoders. Completion sets \(\sigma_{\mathrm{start}}=0\); synthesis samples the perturbed starting state once and keeps the unperturbed evidence endpoint as the network condition throughout integration. \textbf{Supplementary Tab.~\ref{tab:reference-settings}} gives the task-specific stochasticity values.

\refstepcounter{suppmethod}
\subsection*{Supplementary Methods \thesuppmethod. Training and Generation}
\label{supp:training}
\paragraph{Variational Learning}Representation learning uses data and evidence views. Let \(\mathcal L_{\mathrm{rec}}^{(k)}\) denote the normalized reconstruction loss for available variables in group \(k\), averaged over training views and latent samples from the selected subset posterior. It uses the observation distributions defined above. The objective combines reconstruction with regularization of the subset posteriors:
\begin{equation}
\begin{split}
\mathcal L_{\mathrm{VAE}}(\phi,\psi)&=\sum_{k=1}^{K}\mathcal L_{\mathrm{rec}}^{(k)}
+\lambda_{\mathrm{KL}}\mathcal L_{\mathrm{KL}},\\
\mathcal L_{\mathrm{KL}}&=\mathbb E_{(\mathbf x_v,\mathbf m_v)}\!\left[
\frac{1}{|\mathcal S_v|}\sum_{S\in\mathcal S_v}
D_{\mathrm{KL}}\!\left(q_{\phi,S}(\mathbf z\mid\mathbf x_{v,S},\mathbf m_{v,S})\,\|\,p(\mathbf z)\right)\right].
\end{split}
\end{equation}
Here \(\lambda_{\mathrm{KL}}\geq0\) weights the regularizer, and the expectation is over training visits and their views. The objective combines group-normalized reconstruction with component-average KL regularization over the subset posteriors of the mixture-of-products construction \cite{mopoe}.

After representation learning, the encoders and decoders are frozen. The data and evidence endpoints are posterior means formed from all groups available in their respective views. This provides a deterministic representation of each view while retaining the same observation model for both generation tasks.

\paragraph{Reconstruction and Regularization}\label{app:variational-objective}
Balanced selection cycles through valid subset components across training examples. Given a selected component,
\begin{equation}
\mathbf z=\boldsymbol\mu_{S,v}+\boldsymbol\sigma_{S,v}\odot\boldsymbol\epsilon,
\qquad \boldsymbol\epsilon\sim\mathcal N(\mathbf0,\mathbf I).
\end{equation}
Training includes data views and randomly constructed evidence views. Reconstruction scores only the current view. Define \(\operatorname{Avg}(\ell,A)\) as an active-variable average within each visit, followed by an average over visits with any active target:
\begin{equation}
\operatorname{Avg}(\ell,A)=\frac{1}{|\mathcal I_A|}
\sum_{i\in\mathcal I_A}\frac{\sum_j A_{ij}\ell_{ij}}{\sum_j A_{ij}},
\qquad \mathcal I_A=\{i:\sum_j A_{ij}>0\}.
\end{equation}
An empty active set contributes zero. Binary losses use binary cross-entropy and categorical losses use categorical negative log-likelihood. Data views score every categorical variable, including absence; evidence views score selected categories. For numerical groups,
\begin{equation}
\begin{split}
\mathcal L_{\mathrm{pres}}^{(k)}&=\operatorname{Avg}\!\left(
\operatorname{BCE}(\mathbf x_{\mathrm{pres}}^{(k)},\boldsymbol\pi_{\mathrm{pres}}^{(k)}),A_{\mathrm{pres}}^{(k)}\right),\\
\mathcal L_{\mathrm{val}}^{(k)}&=\operatorname{Avg}\!\left(
\frac{(\mathbf x_{\mathrm{val}}^{(k)}-\widehat{\mathbf x}_{\mathrm{val}}^{(k)})^2}{2\sigma_{\mathrm{val}}^2},A_{\mathrm{val}}^{(k)}\right),\\
\mathcal L_{\mathrm{rec}}^{(k)}&=\tfrac12(\mathcal L_{\mathrm{pres}}^{(k)}+\mathcal L_{\mathrm{val}}^{(k)}),
\qquad \mathcal L_{\mathrm{rec}}=\sum_{k=1}^{K}\mathcal L_{\mathrm{rec}}^{(k)}.
\end{split}
\end{equation}
Here \(A_{\mathrm{pres}}^{(k)}\) selects available presence targets and \(A_{\mathrm{val}}^{(k)}\) additionally requires a present measurement. The value loss omits the fixed Gaussian normalization constant. The two numerical branches use separate active-set denominators before their half-weighted combination.

All losses and masks above refer to the current sample and view; the view index is omitted for readability. The presence probabilities \(\boldsymbol\pi_{\mathrm{pres}}^{(k)}\) are obtained from the decoder logits. Reconstruction uses the selected subset sample, whereas the regularizer above averages the component KLs over all valid subsets.

For fixed view weights \(w_S=1/|\mathcal S_v|\), write \(q=\sum_S w_S q_S\), suppressing the view conditions. The distinction between component and mixture regularization follows from
\begin{equation}
\sum_S w_S D_{\mathrm{KL}}(q_S\|p)
=D_{\mathrm{KL}}(q\|p)+\sum_S w_S D_{\mathrm{KL}}(q_S\|q).
\end{equation}
Thus the component average bounds the mixture KL from above.

\paragraph{Training Schedule}
During VAE warmup in the reference implementation, the fraction of evidence views increases linearly from zero to one half, while the evidence-fraction lower bound decreases from the configured upper bound to its minimum. The KL coefficient warms up over the same interval. Thereafter, half the minibatch contains data views and half evidence views. Bridge fitting and generation draw partial evidence under the configured task bounds. Optimization settings are listed in \textbf{Tab.~\ref{tab:reference-settings}}.

\paragraph{Bridge Training}
The endpoint predictor is learned through weighted endpoint regression \cite{ddbm}:
\begin{equation}
\mathcal L_{\mathrm{bridge}}(\theta)
=\mathbb E\!\left[\lambda(t)\operatorname{MSE}
\big(F_\theta(\mathbf z_t,t,\mathbf z_T),\mathbf z_0\big)\right],
\qquad \lambda(t)>0.
\end{equation}
The expectation covers paired visit views, bridge time, intermediate noise, and task-specific endpoint perturbation. With finite second moments, the unrestricted population optimum under this training distribution is \(\mathbb E[\mathbf z_0\mid\mathbf z_t,t,\mathbf z_T]\). Positive time-only weighting does not change that conditional-mean target. The predictor parameterization is specified under \hyperref[app:preconditioning]{Endpoint Preconditioning} in Supplementary Methods~\ref{supp:bridge}.

The bridge objective uses \(\lambda(t)=c_{\mathrm{out}}^{-2}(t)=v_t/r_t\), and MSE averages over latent coordinates. Training samples time uniformly over the configured positive interval together with intermediate noise and the task-specific starting state. Only \(f_\theta\) is learned; representation parameters and analytic coefficients remain fixed. The noise embedding \(c_{\mathrm{noise}}(t)\) is a deterministic transformation of time.

\paragraph{Record Generation}
The frozen decoders map the final shared latent state to the clinical modalities:
\begin{equation}
\widehat{\mathbf x}=\mathcal G\!\left(D_\psi(\mathbf z_0^{\mathrm{gen}})\right).
\label{eq:data-decoder}
\end{equation}
Here \(D_\psi\) produces the observation-model parameters and \(\mathcal G\) applies the output rules to generate a record from them. The rules for each variable type are specified under Output Rules below.

\paragraph{Completion}
Completion uses the target visit's evidence as both the starting state and the neural condition:
\begin{equation}
\mathbf z_T=E_\phi(\mathbf x_e,\mathbf m_e),\qquad
\mathbf z_T^{\mathrm{start}}=\mathbf z_T.
\end{equation}
Stochastic reverse sampling generates variation in the unspecified information. After decoding, the evidence-covering rule of the main text (Eq.~\ref*{eq:covering}) restores the supplied values.

\paragraph{Synthesis}
\label{sec:data-synthesis}
For a reference visit \(\mathbf r\in\mathcal R\) with the requested principal diagnosis,
\begin{equation}
\mathbf r_e=\mathbf m_e\odot\mathbf r,\qquad
\mathbf z_T=E_\phi(\mathbf r_e,\mathbf m_e),\qquad
\mathbf z_T^{\mathrm{start}}=\mathbf z_T+\sigma_{\mathrm{start}}\boldsymbol\xi,
\quad\boldsymbol\xi\sim\mathcal N(\mathbf0,\mathbf I).
\end{equation}
The nonnegative endpoint perturbation scale \(\sigma_{\mathrm{start}}\) controls the Gaussian perturbation applied to the evidence endpoint. The noise \(\boldsymbol\xi\) is independent of the paired visit views and intermediate bridge noise. This separates variation in the starting latent state from the reference evidence that conditions generation \cite{ecsi}: the unperturbed evidence endpoint remains the network condition. The synthesis bridge is trained with the same perturbation rule, with the score approximation specified in Eq.~\eqref{eq:plugin-score}. After reverse sampling and decoding, the synthetic visit is
\begin{equation}
\mathbf x_{\mathrm{synthetic}}=\mathbf d_{\mathrm{diag}}
+(\mathbf1-\mathbf d_{\mathrm{diag}})\odot\widehat{\mathbf x}.
\end{equation}
Only the principal diagnosis is explicitly retained. The other clinical variables are generated jointly from the shared latent state, conditioned on the reference evidence.

\paragraph{Output Rules}
\label{app:output-rules}

Bernoulli sampling draws binary outcomes and categorical states from their respective probabilities. Cali-T fits fixed thresholds for binary and categorical-presence probabilities on decoded training predictions to approximate training-set prevalences, subject to probability ties. The thresholds are reused for generated records; when a categorical variable is present, its most probable non-absence category is selected. Numerical values use decoder estimates bounded according to the data representation, gated by generated measurement presence, rather than Gaussian draws. Both output rules use the same trained observation model.

For completion, supplied evidence is restored after applying the output rules. Completed numerical values are multiplied by completed measurement presence, including when the supplied evidence specifies an absent measurement. The masks obey the presence-value constraints above. In synthesis, only the principal-diagnosis coordinate is explicitly retained. The randomness and calibration settings of \(\mathcal G\) are implicit in Eq.~\eqref{eq:data-decoder}.

\paragraph{Reference Implementation}\label{app:reference-settings}

\textbf{Tab.~\ref{tab:reference-settings}} summarizes the training and sampling settings for the main experiments. The default evidence fraction is drawn uniformly from \([0.3,0.7]\) for completion and \([0.1,0.5]\) for synthesis. The corresponding starting-state perturbation scales are 0 and 0.5.

The variance schedule first constructs a 1,000-element squared linear grid between \(\sqrt{0.1/1000}\) and \(\sqrt{0.3/1000}\). Its first 500 entries are concatenated with their reverse to obtain the symmetric increments in Eq.~\eqref{eq:finite-bridge}. For \(S\) positive sampling times, the reference reverse grid is
\begin{equation}
t_i=\left[(1-u_i)(1-10^{-4})^{1/\rho}
             +u_i(10^{-4})^{1/\rho}\right]^\rho,
\qquad u_i=\frac{i}{S-1},\quad i=0,\ldots,S-1,
\end{equation}
followed by \(t_S=0\). The stochastic and final updates are specified under \hyperref[app:reverse-integration]{Numerical Reverse Integration} in Supplementary Methods~\ref{supp:bridge}.

\refstepcounter{suppmethod}
\subsection*{Supplementary Methods \thesuppmethod. Evaluation Measures and Settings}
\label{supp:evaluation}
\label{sec:evaluation-dimensions}

Predictive utility is a primary evaluation criterion for both tasks, asking whether generated or completed visits support clinical phenotype prediction on real test data; disclosure is an additional primary criterion for synthesis. Diversity and fidelity provide complementary evidence on coverage of the real-data distribution and preservation of feature distributions and dependencies. Each measure is defined together with its evaluation settings below.

\paragraph{Utility}

Utility compares Train on Synthetic, Test on Real (TSTR) with Train on Real, Test on Real (TRTR), following synthetic-EHR benchmarking practice \cite{ehrbenchmark}. For completion, the TSTR arm uses completed training visits. Within each run, LightGBM classifiers use a shared configuration, matched training-set sizes, and the same real test partition \cite{lightgbm}. Diagnosis features that directly define the targets are removed from classifier inputs. The tasks identify recorded phenotypes rather than future outcomes.

MIMIC-IV uses two target sets on the same real test visits. The first comprises ten fixed principal-diagnosis groups; its inputs mask each visit's principal diagnosis and all ten target diagnosis columns. The second comprises seven Phecode-based phenotypes; all diagnosis columns associated with these targets are masked. Both use the remaining diagnosis indicators, procedure and medication indicators, clinical measurement values or categories, and measurement-presence indicators. For the second set, real labels use the original diagnosis codes and Phecode mapping, whereas synthetic labels use mapped three-character diagnosis groups. Each Duke representation uses ten fixed diagnosis targets; inputs mask each visit's assigned principal diagnosis and all ten target columns and include the remaining diagnosis indicators, medication indicators, and clinical measurements with presence indicators. The first and second MIMIC-IV metric pairs in the Utility tables correspond to the ten- and seven-target sets, respectively.

For target \(j\), the signed gaps are
\begin{equation}
\begin{aligned}
\Delta\mathrm{AUROC}_j
&=\mathrm{AUROC}_{\mathrm{TRTR},j}
 -\mathrm{AUROC}_{\mathrm{TSTR},j},\\
\Delta\mathrm{AUPRC}_j
&=\mathrm{AUPRC}_{\mathrm{TRTR},j}
 -\mathrm{AUPRC}_{\mathrm{TSTR},j}.
\end{aligned}
\end{equation}
Here, AUPRC is computed as average precision, the recall-increment-weighted sum of precision, rather than trapezoidal integration. For \(J\) valid targets, each run's reported gap is the arithmetic mean of its \(J\) target-specific gaps. The main comparisons and evidence-fraction analyses use the seed-42 target averages; component analyses additionally average these gaps over seeds 42, 123, and 2026. Zero denotes parity, positive gaps a TSTR shortfall, and negative gaps higher TSTR performance. The comparison uses closeness to zero after the applicable averaging, retaining the sign in the table.

Each binary LightGBM classifier has 200 estimators, learning rate 0.05, 31 leaves, \texttt{subsample=0.9}, and \texttt{colsample\_bytree=0.9}. Within each run, the real-trained and generated-data-trained classifiers share these hyperparameters and the same classifier random seed. A single-class training target receives its observed prevalence as a constant score.

All figures use seed 42. Utility curves use saved predictions from that run. MoPoE-VAE and medGAN classifiers were refitted for curve export; the maximum macro-score differences from the tabulated seed-42 reports were below \(0.00015\) for AUROC and \(0.000024\) for AUPRC.

\paragraph{Diversity}

Real and generated visits use the feature representation specified for BCMD below and are embedded in the same PCA space fitted on the real training partition. Average Pairwise Distance Ratio (APDR) compares global dispersion. For equal-sized real and generated sets \(R=\{\mathbf r_i\}_{i=1}^{N}\) and \(S=\{\mathbf s_i\}_{i=1}^{N}\), let \(\mathcal Q\) be a common collection of sampled index pairs \((a,b)\), with \(a\ne b\). The same pairs are used for both samples:
\begin{equation}
\mathrm{APDR}=
\frac{\sum_{(a,b)\in\mathcal Q}\|\mathbf s_a-\mathbf s_b\|_2}
{\sum_{(a,b)\in\mathcal Q}\|\mathbf r_a-\mathbf r_b\|_2}.
\end{equation}
The same normalization cancels from numerator and denominator. A value near one indicates similar average dispersion; smaller values indicate contraction and larger values overdispersion.

Coverage adapts the local-support measure of Naeem et al.~\cite{densitycoverage}. If \(\rho_i\) is the distance from \(\mathbf r_i\) to its \(k\)-th nearest non-self real neighbor, where \(k\) specifies the neighborhood size,
\begin{equation}
\mathrm{Coverage}=\frac{1}{N}\sum_{i=1}^{N}
\mathbb I\!\left[\min_j\|\mathbf r_i-\mathbf s_j\|_2\leq\rho_i\right].
\end{equation}
Higher Coverage means that more real-data neighborhoods contain a generated sample. APDR and Coverage are interpreted together because global spread does not determine local support.

The common PCA space has 32 dimensions. For APDR, \(\mathcal Q\) is a multiset of \(N\) ordered index pairs sampled with replacement, excluding self-pairs, where \(N\) is the size of each matched sample. Coverage uses the fifth-nearest non-self real neighbor, corresponding to \(k=5\).

\paragraph{Fidelity}
\label{sec:evaluation-fidelity}

Feature Distribution Distance (FDD) measures marginal discrepancies. For a binary or non-absence one-hot feature \(j\), the contribution is the absolute prevalence difference,
\begin{equation}
d_j=\left|p_{\mathrm{real},j}-p_{\mathrm{gen},j}\right|.
\end{equation}
For a standardized numerical variable, presence and present-value discrepancies are
\begin{equation}
\begin{split}
d_{\mathrm{pres},j}
&=\left|p^{\mathrm{pres}}_{\mathrm{real},j}
       -p^{\mathrm{pres}}_{\mathrm{gen},j}\right|,\\
d_{\mathrm{val},j}
&=\frac{1}{L}\sum_{\ell=1}^{L}
\left|Q_{\mathrm{real},j}(u_\ell)
      -Q_{\mathrm{gen},j}(u_\ell)\right| .
\end{split}
\end{equation}
Here, \(L\) is the number of quantile levels. Quantiles use present values only, on the grid \(u_\ell=(\ell-1)/(L-1)\). A numerical variable contributes \(d_j=(d_{\mathrm{pres},j}+d_{\mathrm{val},j})/2\). Here \(D\) counts one term per binary feature or non-absence categorical indicator and one combined term per numerical variable; numerical presence is not counted again:
\begin{equation}
\mathrm{FDD}=\frac{1}{D}\sum_{j=1}^{D}d_j .
\end{equation}
The quantile-grid term approximates a one-dimensional first-order Wasserstein distance. Because measurement representation changes the terms and their count, FDD comparisons are interpreted within a data setting.

Blockwise Correlation Matrix Distance (BCMD) measures dependence. The evaluation representation uses zero-filled feature values. MIMIC-IV appends clinical measurement-presence indicators, and DH includes numerical-presence indicators; DH-D uses non-absence categorical indicators without an additional presence column. Let \(B\) be the number of clinical modalities, four for MIMIC-IV and three for the Duke Health data, and let \(D_a\) be the width of modality \(a\). Using Pearson correlation matrices,
\begin{equation}
d_{a,b}=
\frac{\left\|C_{\mathrm{real}}^{(a,b)}
             -C_{\mathrm{gen}}^{(a,b)}\right\|_F}
{\sqrt{D_aD_b}},
\qquad
\mathrm{BCMD}=\frac{2}{B(B+1)}
\sum_{a\leq b}d_{a,b}.
\end{equation}
Within-modality blocks retain their diagonals. These clinical modalities may contain multiple modeling groups used for variational fusion. BCMD gives equal weight to each within- or cross-modality block. Lower FDD and BCMD indicate smaller discrepancies; both are study-defined summaries.

Numerical FDD uses \(L=100\) equally spaced quantile levels, including both endpoints. The present-value term is zero if both present-value sets are empty and one if exactly one is empty. For BCMD, correlations involving a column with sample standard deviation at most \(10^{-8}\) are set to zero; other diagonal entries are one.

\paragraph{Disclosure risk}

Disclosure risk is assessed using a membership inference attack (MIA) and an attribute inference attack (AIA) \cite{shokri2017membership,stadler2022synthetic}. The evaluations apply distance-based attacks used in synthetic-EHR studies \cite{medgan,emrwgan} to the released synthetic training visits.

For MIA, the attacker scores a query visit by the negative distance to its nearest synthetic training visit. The mixed distance combines Jaccard discrepancies for binary clinical modalities with categorical mismatch or empirical-quantile differences for clinical measurements, giving clinical modalities equal weight. The attack compares patient-disjoint training members and test nonmembers, with at most 5,000 patients per group and one visit per patient. Attack performance is measured by AUROC, with members as the positive class. A score near 0.5 indicates little discrimination under this attack; either direction of departure can carry membership information.

For AIA, the attacker observes a randomly selected half of the original variables in each modality, together with the principal diagnosis, and predicts hidden diagnosis and medication codes using five nearest neighbors. Eligible targets have at least 20 positive examples in an independent real auxiliary set and contain both classes among the hidden-target queries. The auxiliary set is drawn from a disjoint 20\% subset of test patients and capped at 5,000 visits; a synthetic reference set of equal size is used for comparison. The auxiliary patients are excluded from the nonmember evaluation pool. Numerical distance transforms are fitted on this auxiliary set. Let \(\mathcal J_{\mathrm{hid}}\) be the valid hidden targets among training-member queries. The reported AUPRC advantage is
\begin{equation}
\Delta\mathrm{AIA}=
\frac{1}{|\mathcal J_{\mathrm{hid}}|}
\sum_{j\in\mathcal J_{\mathrm{hid}}}
\left(\mathrm{AUPRC}_{\mathrm{syn},j}
      -\mathrm{AUPRC}_{\mathrm{aux},j}\right).
\end{equation}
A positive value indicates improved inference from the synthetic reference under this control; a negative value indicates worse inference. \(\Delta\mathrm{AIA}\) compares synthetic-reference and real-auxiliary-reference inference on training-member queries. The two attacks assess disclosure risk under the specified adversarial knowledge and reference-set protocols.

\paragraph{Evidence-Fraction Configurations}
For the MIMIC-IV synthesis analysis, models were trained and evaluated under twelve evidence-fraction intervals: \([0.1,0.3]\), \([0.1,0.4]\), \([0.1,0.5]\), \([0.1,0.6]\), \([0.1,0.7]\), \([0.2,0.4]\), \([0.2,0.5]\), \([0.2,0.6]\), \([0.2,0.7]\), \([0.3,0.5]\), \([0.3,0.6]\), and \([0.3,0.7]\). All configurations used the same patient partitions, model architecture, training budget, and other experimental settings. Each interval was summarized by its mean evidence fraction. The resulting configuration-level measures are reported in \textbf{Supplementary Tabs.~\ref{tab:robustness-fidelity}-\ref{tab:robustness-privacy}}.

\refstepcounter{suppmethod}
\subsection*{Supplementary Methods \thesuppmethod. Component Analysis Design}
\label{supp:components}
{\color{blue}

\textcolor{black}{Component analyses were conducted for MIMIC-IV synthesis, varying one component at a time with the remaining settings fixed within each study. Entries are arithmetic means over seeds 42, 123, and 2026, without standard deviations. FDD and BCMD are reported on the \(\times10^{3}\) scale; mean utility gaps are assessed by absolute distance from zero, Coverage is maximized, and mean APDR is assessed by distance from one.}

\color{black}
Patient partitions are regenerated for each seed. MIMIC-IV training, validation, and test visit counts are 196,188, 24,292, and 24,650 for seed 42; 196,263, 24,036, and 24,831 for seed 123; and 196,295, 24,490, and 24,345 for seed 2026. Within a seed, all compared component configurations share the same patient partitions and real-trained utility reference.

\paragraph{Availability Encoding}
We compare no explicit availability encoding, availability concatenation, a learned availability embedding, and availability gating. The gate modulates the first hidden layer without changing its width and starts from the identity because its affine parameters are zero-initialized.

\paragraph{KL Regularization}
We vary the KL weight over \(10^{-6},10^{-5},10^{-4},10^{-3},10^{-2},10^{-1}\), holding the remaining components fixed.

\paragraph{Forward Bridge}
We compare variance-exploding (VE), variance-preserving (VP), Linear, and Schr\"odinger paths within the same endpoint-prediction framework.

\paragraph{Synthesis Sampling}
The sampler analysis separates perturbation of the evidence-endpoint distribution from stochasticity along the reverse path. Hybrid SDE-ODE sampling crosses evidence-endpoint perturbation levels \(\{0,0.5,1\}\) with stochasticity levels \(\{0,0.125,0.25,0.33,0.5\}\). Stochastic-interpolant sampling crosses the same endpoint-perturbation levels with stochasticity levels \(\{0,0.3,0.5,0.8,1\}\). The two reported grids contain fifteen configurations each. At zero endpoint perturbation, the unperturbed evidence endpoint initializes synthesis, and the decoded record retains only the principal diagnosis.

Within each seed, the sampling study shares one trained variational representation and trains a separate bridge for each endpoint-perturbation level. At a fixed level, sampler variants use the same bridge checkpoint, partitions, and evaluation protocol. Hybrid sampling combines stochastic Euler steps with Heun updates of the deterministic substep and a terminal Euler update. The stochastic-interpolant implementation uses Euler-Maruyama updates followed by coefficient-based terminal transfers. The samplers differ in stochastic dynamics, integration rules, and neural-evaluation counts.

The synthesis configuration used in the main comparisons has KL weight \(10^{-5}\), availability gating, the Schr\"odinger bridge, evidence-endpoint perturbation 0.5, and stochasticity 0.25. The completion configuration uses an unperturbed endpoint and stochasticity 0.5.
}

\clearpage

\section*{Supplementary Results}
\subsection*{Study Populations and Predictive Utility}
\label{supp:results-population}

The main experiments used patient-level partitions into training, validation, and test sets, comprising 196,188, 24,292, and 24,650 visits in MIMIC-IV and 750,817, 92,220, and 93,434 visits in Duke Health, respectively. The MIMIC-IV partitions used for the component analyses are described in Supplementary Methods~\ref{supp:components}.

\textbf{Supplementary Tab.~\ref{tab:population}} provides characteristics by data partition. \textbf{Supplementary Figs.~\ref{fig:utility-b}-\ref{fig:diagnosis-b}} show the additional MIMIC-IV utility curves.

\subsection*{Evidence-Fraction Configurations}
\label{supp:results-evidence}

\textbf{Tabs.~\ref{tab:robustness-fidelity}-\ref{tab:robustness-privacy}} report MIMIC-IV synthesis with EHRDiB under the twelve evidence-fraction intervals described in Supplementary Methods~\ref{supp:evaluation}.

\subsection*{Component Analyses}
\label{supp:results-components}
{

\paragraph{Availability Mask Encoding}

\textbf{Tab.~\ref{tab:availability}} shows that gating gives the lowest FDD and BCMD, highest Coverage, and smallest AUROC and AUPRC gaps among the four included encodings. Compared with no explicit availability encoding, gating reduced FDD from 23.11 to 22.11 and BCMD from 15.19 to 14.94, while increasing Coverage and narrowing utility gaps, supporting the use of availability gating. Availability concatenation gives APDR closest to one.

\paragraph{KL Regularization}

\textbf{Tab.~\ref{tab:kl}} shows a stable low-weight region and marked degradation at larger weights. The selected \(10^{-5}\) setting has the lowest FDD and BCMD, APDR closest to one, and the smallest AUROC and AUPRC gaps. The \(10^{-4}\) setting gives the highest Coverage. These results support the selected weight for the evaluated architecture.

\paragraph{Forward Bridge}

\textbf{Tab.~\ref{tab:bridge}} shows that the Schr\"odinger path gives the lowest FDD and BCMD, APDR closest to one, and the smallest phenotyping AUROC gap. Linear gives the highest Coverage and smallest AUPRC gap. The selected path therefore offers a favorable combination of the measured outcomes, with different paths leading different measures.

\paragraph{Synthesis Sampling}

\textbf{Tabs.~\ref{tab:sam-hybrid} and~\ref{tab:sam-interpolant}} show that the best results for different measures occur under different combinations of sampler and hyperparameters. Across all tested configurations, the lowest FDD and APDR closest to one occurred with the hybrid SDE-ODE sampler at endpoint perturbation 0 and stochasticity 0.125; the lowest BCMD occurred with the same sampler at endpoint perturbation 0 and stochasticity 0. The highest Coverage occurred with the stochastic-interpolant sampler at endpoint perturbation 1 and stochasticity 0, whereas the smallest AUROC and AUPRC gaps occurred with the hybrid SDE-ODE sampler at the same hyperparameter values. These results support jointly selecting the sampler, endpoint perturbation, and stochasticity according to the specific experimental configuration and evaluation priorities.

}

\clearpage
\section*{Supplementary Tables}
\label{supp:tables}
\begingroup
\makeatletter
\setlength{\@fptop}{0pt}
\setlength{\@fpsep}{14pt}
\setlength{\@fpbot}{0pt plus 1fil}
\makeatother
\input{tables/population}
\FloatBarrier

\input{tables/robustness_fidelity}
\FloatBarrier

\input{tables/robustness_diversity}
\FloatBarrier

\input{tables/robustness_utility}
\FloatBarrier

\input{tables/robustness_privacy}
\FloatBarrier

\input{tables/availability}
\FloatBarrier

\input{tables/kl_regularization}
\FloatBarrier

\input{tables/forward_bridge}
\FloatBarrier

\input{tables/sampling_hybrid}
\FloatBarrier

\input{tables/sampling_interpolant}
\FloatBarrier

\input{tables/reference_settings}
\FloatBarrier

\endgroup
\clearpage
\section*{Supplementary Figures}
\label{supp:figures}
\clearpage
\clearpage

%% file: tables/population.tex
\begin{table}[H]
\centering
\caption{Characteristics of the study populations by data partition. Age is reported as median (IQR), estimated at the first included admission for MIMIC-IV and measured at the PREVENT study baseline for Duke Health. Age and sex summarize unique patients; other distributions summarize visits. Measurement presence uses 462 distinct laboratory items in MIMIC-IV and eight clinical indicators in Duke Health. Target prevalence spans ten principal-diagnosis targets in MIMIC-IV and ten diagnosis targets in Duke Health within each partition. DH and DH-D share the same patients and visits.}
\label{tab:population}
\small
\begin{tabularx}{\linewidth}{@{}>{\raggedright\arraybackslash}Xrrrrrr@{}}
\toprule
 & \multicolumn{3}{c}{MIMIC-IV} & \multicolumn{3}{c}{Duke Health} \\
\cmidrule(lr){2-4}\cmidrule(l){5-7}
Characteristic & Training & Validation & Test & Training & Validation & Test \\
\midrule
Patients, n & 95,532 & 11,946 & 11,947 & 96,704 & 12,073 & 12,079 \\
Visits, n & 196,188 & 24,292 & 24,650 & 750,817 & 92,220 & 93,434 \\
Age, years & \begin{tabular}[t]{@{}r@{}}61\\(43-74)\end{tabular} & \begin{tabular}[t]{@{}r@{}}61\\(43-74)\end{tabular} & \begin{tabular}[t]{@{}r@{}}61\\(44-74)\end{tabular} & \begin{tabular}[t]{@{}r@{}}53.9\\(43.7-63.8)\end{tabular} & \begin{tabular}[t]{@{}r@{}}53.9\\(43.6-63.8)\end{tabular} & \begin{tabular}[t]{@{}r@{}}54.0\\(43.9-63.8)\end{tabular} \\
Female, \% & 52.5 & 52.9 & 52.8 & 56.7 & 57.0 & 57.3 \\
Measurement variables present per visit, median (IQR), \% & \begin{tabular}[t]{@{}r@{}}7.8\\(5.2-11.3)\end{tabular} & \begin{tabular}[t]{@{}r@{}}7.8\\(4.9-11.3)\end{tabular} & \begin{tabular}[t]{@{}r@{}}7.6\\(5.0-11.3)\end{tabular} & \begin{tabular}[t]{@{}r@{}}25.0\\(25.0-37.5)\end{tabular} & \begin{tabular}[t]{@{}r@{}}25.0\\(25.0-37.5)\end{tabular} & \begin{tabular}[t]{@{}r@{}}25.0\\(25.0-37.5)\end{tabular} \\
Diagnosis codes per visit, median (IQR) & \begin{tabular}[t]{@{}r@{}}11\\(7-17)\end{tabular} & \begin{tabular}[t]{@{}r@{}}11\\(7-17)\end{tabular} & \begin{tabular}[t]{@{}r@{}}11\\(7-17)\end{tabular} & \begin{tabular}[t]{@{}r@{}}3\\(2-5)\end{tabular} & \begin{tabular}[t]{@{}r@{}}3\\(2-5)\end{tabular} & \begin{tabular}[t]{@{}r@{}}3\\(2-5)\end{tabular} \\
Medications per visit, median (IQR) & \begin{tabular}[t]{@{}r@{}}14\\(8-22)\end{tabular} & \begin{tabular}[t]{@{}r@{}}14\\(8-22)\end{tabular} & \begin{tabular}[t]{@{}r@{}}14\\(8-22)\end{tabular} & \begin{tabular}[t]{@{}r@{}}2\\(1-4)\end{tabular} & \begin{tabular}[t]{@{}r@{}}2\\(1-4)\end{tabular} & \begin{tabular}[t]{@{}r@{}}2\\(1-4)\end{tabular} \\
Target prevalence, range, \% & 0.5-3.5 & 0.5-3.4 & 0.5-3.3 & 1.8-27.8 & 1.8-28.3 & 1.7-27.6 \\
\bottomrule
\end{tabularx}
\end{table}

%% file: tables/robustness_fidelity.tex
\begin{table}[H]
\centering
\caption{Distributional fidelity of EHRDiB in MIMIC-IV synthesis across evidence-fraction intervals. FDD, feature distribution distance; BCMD, blockwise correlation matrix distance. Bold indicates the lowest value for each metric.}
\label{tab:robustness-fidelity}
\begingroup\fontsize{9}{11}\selectfont
\setlength{\tabcolsep}{3pt}\renewcommand{\arraystretch}{1.05}
\begin{tabular}{@{}l@{\hspace{16pt}}r@{\hspace{12pt}}r@{}}
\toprule
 & \multicolumn{2}{c}{Fidelity (\(\times10^{3}\))} \\
\cmidrule(l){2-3}
Evidence-fraction interval & FDD & BCMD \\
\midrule
\([0.1,0.3]\) & 24.01 & 15.43 \\
\([0.1,0.4]\) & 22.92 & 15.05 \\
\([0.1,0.5]\) & 22.38 & 14.72 \\
\([0.1,0.6]\) & 23.21 & 14.74 \\
\([0.1,0.7]\) & 21.87 & 14.20 \\
\([0.2,0.4]\) & 22.42 & 14.96 \\
\([0.2,0.5]\) & 22.19 & 14.50 \\
\([0.2,0.6]\) & 21.50 & 14.34 \\
\([0.2,0.7]\) & 20.86 & 14.12 \\
\([0.3,0.5]\) & 21.82 & 14.27 \\
\([0.3,0.6]\) & \textbf{20.84} & 14.14 \\
\([0.3,0.7]\) & 21.19 & \textbf{13.70} \\
\bottomrule
\end{tabular}
\endgroup

\end{table}

%% file: tables/robustness_diversity.tex
\begin{table}[H]
\centering
\caption{Diversity of EHRDiB in MIMIC-IV synthesis across evidence-fraction intervals. Coverage measures the proportion of real-data neighborhoods containing a synthetic record in the shared principal-component space; the average pairwise distance ratio (APDR) compares the dispersion of synthetic and real records. Bold indicates the highest Coverage and APDR closest to 1.}
\label{tab:robustness-diversity}
\begingroup\fontsize{9}{11}\selectfont
\setlength{\tabcolsep}{3pt}\renewcommand{\arraystretch}{1.05}
\begin{tabular}{@{}l@{\hspace{16pt}}r@{\hspace{12pt}}r@{}}
\toprule
Evidence-fraction interval & \multicolumn{1}{c}{APDR} & \multicolumn{1}{c}{Coverage} \\
\midrule
\([0.1,0.3]\) & 0.9407 & 0.9759 \\
\([0.1,0.4]\) & 0.9471 & 0.9786 \\
\([0.1,0.5]\) & 0.9548 & 0.9815 \\
\([0.1,0.6]\) & 0.9583 & 0.9865 \\
\([0.1,0.7]\) & 0.9621 & 0.9883 \\
\([0.2,0.4]\) & 0.9587 & 0.9830 \\
\([0.2,0.5]\) & 0.9588 & 0.9852 \\
\([0.2,0.6]\) & 0.9631 & 0.9883 \\
\([0.2,0.7]\) & 0.9659 & 0.9905 \\
\([0.3,0.5]\) & 0.9695 & 0.9885 \\
\([0.3,0.6]\) & \textbf{0.9724} & 0.9920 \\
\([0.3,0.7]\) & 0.9632 & \textbf{0.9933} \\
\bottomrule
\end{tabular}
\endgroup

\end{table}

%% file: tables/robustness_utility.tex
\begin{table}[H]
\centering
\caption{Predictive utility of EHRDiB in MIMIC-IV synthesis across evidence-fraction intervals. Utility gaps are differences in AUROC and AUPRC between LightGBM classifiers trained on real and synthetic records. Set 1 comprises ten principal-diagnosis targets; set 2 comprises seven Phecode-defined phenotypes. Bold indicates the smallest utility gap for each metric and target set.}
\label{tab:robustness-utility}
\begingroup\fontsize{9}{11}\selectfont
\setlength{\tabcolsep}{3pt}\renewcommand{\arraystretch}{1.05}
\begin{tabular}{@{}l@{\hspace{16pt}}r@{\hspace{12pt}}r@{\hspace{12pt}}r@{\hspace{12pt}}r@{}}
\toprule
 & \multicolumn{2}{c}{Utility, set 1} & \multicolumn{2}{c}{Utility, set 2} \\
\cmidrule(lr){2-3}\cmidrule(l){4-5}
Evidence-fraction interval & \multicolumn{1}{c}{\(\Delta\)AUROC} & \multicolumn{1}{c}{\(\Delta\)AUPRC} & \multicolumn{1}{c}{\(\Delta\)AUROC} & \multicolumn{1}{c}{\(\Delta\)AUPRC} \\
\midrule
\([0.1,0.3]\) & 0.0054 & 0.0625 & 0.0426 & 0.1018 \\
\([0.1,0.4]\) & 0.0043 & 0.0493 & 0.0396 & 0.0996 \\
\([0.1,0.5]\) & 0.0042 & 0.0457 & 0.0410 & 0.0969 \\
\([0.1,0.6]\) & 0.0037 & 0.0471 & 0.0391 & 0.1005 \\
\([0.1,0.7]\) & 0.0034 & 0.0450 & 0.0382 & 0.0970 \\
\([0.2,0.4]\) & 0.0033 & 0.0404 & 0.0396 & 0.0995 \\
\([0.2,0.5]\) & 0.0049 & \textbf{0.0327} & 0.0364 & 0.0959 \\
\([0.2,0.6]\) & 0.0040 & 0.0385 & 0.0353 & 0.0954 \\
\([0.2,0.7]\) & 0.0035 & 0.0331 & 0.0363 & \textbf{0.0914} \\
\([0.3,0.5]\) & 0.0035 & 0.0412 & \textbf{0.0342} & 0.0962 \\
\([0.3,0.6]\) & 0.0037 & 0.0463 & 0.0345 & 0.0915 \\
\([0.3,0.7]\) & \textbf{0.0027} & 0.0432 & 0.0347 & 0.0981 \\
\bottomrule
\end{tabular}
\endgroup

\end{table}

%% file: tables/robustness_privacy.tex
\begin{table}[H]
\centering
\caption{Disclosure risk of EHRDiB in MIMIC-IV synthesis across evidence-fraction intervals. Membership inference attack (MIA) performance is measured by AUROC, with chance at 0.5. Attribute inference advantage (\(\Delta\)AIA) is the difference in attack AUPRC between a synthetic reference and a real auxiliary reference. Bold indicates MIA AUROC closest to 0.5 and the lowest \(\Delta\)AIA.}
\label{tab:robustness-privacy}
\begingroup\fontsize{9}{11}\selectfont
\setlength{\tabcolsep}{3pt}\renewcommand{\arraystretch}{1.05}
\begin{tabular}{@{}l@{\hspace{16pt}}r@{\hspace{12pt}}r@{}}
\toprule
Evidence-fraction interval & \multicolumn{1}{c}{MIA} & \multicolumn{1}{c}{\(\Delta\)AIA} \\
\midrule
\([0.1,0.3]\) & \textbf{0.5162} & -0.0036 \\
\([0.1,0.4]\) & 0.5218 & -0.0028 \\
\([0.1,0.5]\) & 0.5336 & \textbf{-0.0049} \\
\([0.1,0.6]\) & 0.5467 & -0.0024 \\
\([0.1,0.7]\) & 0.5694 & -0.0035 \\
\([0.2,0.4]\) & 0.5273 & -0.0023 \\
\([0.2,0.5]\) & 0.5431 & -0.0037 \\
\([0.2,0.6]\) & 0.5593 & -0.0034 \\
\([0.2,0.7]\) & 0.5809 & -0.0048 \\
\([0.3,0.5]\) & 0.5582 & -0.0043 \\
\([0.3,0.6]\) & 0.5744 & -0.0024 \\
\([0.3,0.7]\) & 0.5984 & -0.0028 \\
\bottomrule
\end{tabular}
\endgroup

\end{table}

%% file: tables/availability.tex
\begin{table}[H]
\centering
\caption{Availability-encoding results for EHRDiB in MIMIC-IV synthesis. Utility gaps are differences in AUROC and AUPRC between LightGBM classifiers trained on real and synthetic records for the ten principal-diagnosis targets. Bold indicates the best result for each metric before rounding.}
\label{tab:availability}
\footnotesize
\setlength{\tabcolsep}{4pt}
\begin{tabular}{@{}lcccccc@{}}
\toprule
 & \multicolumn{2}{c}{Fidelity (\(\times10^{3}\))} & \multicolumn{2}{c}{Diversity} & \multicolumn{2}{c}{Utility} \\
\cmidrule(lr){2-3}\cmidrule(lr){4-5}\cmidrule(l){6-7}
Encoding & FDD & BCMD & Coverage & APDR & \(\Delta\)AUROC & \(\Delta\)AUPRC \\
\midrule
No explicit encoding & 23.11 & 15.19 & 0.986 & 0.958 & 0.006 & 0.069 \\
Availability concatenation & 22.91 & 15.41 & 0.985 & \textbf{0.958} & 0.007 & 0.074 \\
Availability embedding & 23.34 & 15.18 & 0.987 & 0.957 & 0.006 & 0.060 \\
Availability gating & \textbf{22.11} & \textbf{14.94} & \textbf{0.988} & 0.958 & \textbf{0.006} & \textbf{0.060} \\
\bottomrule
\end{tabular}
\end{table}

%% file: tables/kl_regularization.tex
\begin{table}[H]
\centering
\caption{KL-regularization results for EHRDiB in MIMIC-IV synthesis. Utility gaps are differences in AUROC and AUPRC between LightGBM classifiers trained on real and synthetic records for the ten principal-diagnosis targets. Bold indicates the best result for each metric before rounding.}
\label{tab:kl}
\footnotesize
\setlength{\tabcolsep}{4pt}
\begin{tabular}{@{}lcccccc@{}}
\toprule
 & \multicolumn{2}{c}{Fidelity (\(\times10^{3}\))} & \multicolumn{2}{c}{Diversity} & \multicolumn{2}{c}{Utility} \\
\cmidrule(lr){2-3}\cmidrule(lr){4-5}\cmidrule(l){6-7}
\(\lambda_{\mathrm{KL}}\) & FDD & BCMD & Coverage & APDR & \(\Delta\)AUROC & \(\Delta\)AUPRC \\
\midrule
\(10^{-1}\) & 44.24 & 29.69 & 0.113 & 0.336 & 0.361 & 0.584 \\
\(10^{-2}\) & 27.37 & 20.36 & 0.895 & 0.849 & 0.084 & 0.393 \\
\(10^{-3}\) & 23.26 & 15.52 & 0.987 & 0.942 & 0.010 & 0.107 \\
\(10^{-4}\) & 22.35 & 15.11 & \textbf{0.989} & 0.952 & 0.006 & 0.068 \\
\(10^{-5}\) & \textbf{22.11} & \textbf{14.94} & 0.988 & \textbf{0.958} & \textbf{0.006} & \textbf{0.060} \\
\(10^{-6}\) & 22.82 & 15.06 & 0.989 & 0.955 & 0.006 & 0.062 \\
\bottomrule
\end{tabular}
\end{table}

%% file: tables/forward_bridge.tex
\begin{table}[H]
\centering
\caption{Forward-bridge results for EHRDiB in MIMIC-IV synthesis. All paths use the same endpoint-prediction framework. Utility gaps are differences in AUROC and AUPRC between LightGBM classifiers trained on real and synthetic records for the ten principal-diagnosis targets. Bold indicates the best result for each metric before rounding.}
\label{tab:bridge}
\footnotesize
\setlength{\tabcolsep}{4pt}
\begin{tabular}{@{}lcccccc@{}}
\toprule
 & \multicolumn{2}{c}{Fidelity (\(\times10^{3}\))} & \multicolumn{2}{c}{Diversity} & \multicolumn{2}{c}{Utility} \\
\cmidrule(lr){2-3}\cmidrule(lr){4-5}\cmidrule(l){6-7}
Bridge & FDD & BCMD & Coverage & APDR & \(\Delta\)AUROC & \(\Delta\)AUPRC \\
\midrule
VE & 22.06 & 14.94 & 0.989 & 0.959 & 0.005 & 0.060 \\
VP & 22.11 & 14.95 & 0.989 & 0.969 & 0.005 & 0.059 \\
Linear & 21.78 & 14.63 & \textbf{0.990} & 0.964 & 0.006 & \textbf{0.051} \\
Schr\"odinger & \textbf{20.91} & \textbf{14.37} & 0.989 & \textbf{0.974} & \textbf{0.005} & 0.052 \\
\bottomrule
\end{tabular}
\end{table}

%% file: tables/sampling_hybrid.tex
\begin{table}[H]
\centering
\caption{Hybrid SDE-ODE sampling results for EHRDiB in MIMIC-IV synthesis, varying evidence-endpoint perturbation and stochasticity. Utility gaps are differences in AUROC and AUPRC between LightGBM classifiers trained on real and synthetic records for the ten principal-diagnosis targets. Bold indicates the best result for each metric before rounding.}
\label{tab:sam-hybrid}
\footnotesize
\setlength{\tabcolsep}{4pt}
\begin{tabular}{@{}lcccccc@{}}
\toprule
 & \multicolumn{2}{c}{Fidelity (\(\times10^{3}\))} & \multicolumn{2}{c}{Diversity} & \multicolumn{2}{c}{Utility} \\
\cmidrule(lr){2-3}\cmidrule(lr){4-5}\cmidrule(l){6-7}
Perturbation, stochasticity & FDD & BCMD & Coverage & APDR & \(\Delta\)AUROC & \(\Delta\)AUPRC \\
\midrule
0, 0 & 23.22 & \textbf{14.22} & 0.990 & 0.921 & 0.004 & 0.053 \\
0, 0.125 & \textbf{20.85} & 14.41 & 0.988 & \textbf{0.986} & 0.005 & 0.049 \\
0, 0.25 & 20.94 & 14.35 & 0.989 & 0.978 & 0.005 & 0.049 \\
0, 0.33 & 20.93 & 14.34 & 0.989 & 0.976 & 0.005 & 0.050 \\
0, 0.5 & 20.95 & 14.36 & 0.989 & 0.972 & 0.005 & 0.048 \\
0.5, 0 & 21.83 & 14.26 & \textbf{0.994} & 0.957 & 0.005 & 0.046 \\
0.5, 0.125 & 21.30 & 14.36 & 0.992 & 0.971 & 0.005 & 0.054 \\
0.5, 0.25 & 21.38 & 14.39 & 0.992 & 0.971 & 0.005 & 0.055 \\
0.5, 0.33 & 21.33 & 14.37 & 0.991 & 0.971 & 0.005 & 0.054 \\
0.5, 0.5 & 21.28 & 14.37 & 0.991 & 0.972 & 0.005 & 0.052 \\
1, 0 & 21.67 & 14.30 & 0.993 & 0.962 & \textbf{0.004} & \textbf{0.046} \\
1, 0.125 & 21.41 & 14.35 & 0.992 & 0.968 & 0.004 & 0.053 \\
1, 0.25 & 21.41 & 14.38 & 0.992 & 0.968 & 0.005 & 0.053 \\
1, 0.33 & 21.44 & 14.40 & 0.991 & 0.969 & 0.005 & 0.050 \\
1, 0.5 & 21.43 & 14.40 & 0.991 & 0.969 & 0.005 & 0.055 \\
\bottomrule
\end{tabular}
\end{table}

%% file: tables/sampling_interpolant.tex
\begin{table}[H]
\centering
\caption{Stochastic-interpolant sampling results for EHRDiB in MIMIC-IV synthesis, varying evidence-endpoint perturbation and stochasticity. Utility gaps are differences in AUROC and AUPRC between LightGBM classifiers trained on real and synthetic records for the ten principal-diagnosis targets. Bold indicates the best result for each metric before rounding.}
\label{tab:sam-interpolant}
\footnotesize
\setlength{\tabcolsep}{4pt}
\begin{tabular}{@{}lcccccc@{}}
\toprule
 & \multicolumn{2}{c}{Fidelity (\(\times10^{3}\))} & \multicolumn{2}{c}{Diversity} & \multicolumn{2}{c}{Utility} \\
\cmidrule(lr){2-3}\cmidrule(lr){4-5}\cmidrule(l){6-7}
Perturbation, stochasticity & FDD & BCMD & Coverage & APDR & \(\Delta\)AUROC & \(\Delta\)AUPRC \\
\midrule
0, 0 & 25.50 & 14.56 & 0.989 & 0.873 & 0.005 & 0.063 \\
0, 0.3 & 22.81 & 14.53 & 0.991 & 0.919 & 0.005 & 0.051 \\
0, 0.5 & 22.03 & 14.51 & 0.991 & 0.936 & 0.005 & 0.050 \\
0, 0.8 & 21.57 & 14.47 & 0.991 & 0.951 & 0.005 & 0.049 \\
0, 1 & 21.40 & 14.44 & 0.990 & 0.956 & 0.005 & 0.052 \\
0.5, 0 & 23.33 & 14.39 & 0.994 & 0.923 & 0.005 & 0.053 \\
0.5, 0.3 & 21.92 & 14.38 & 0.993 & 0.947 & 0.005 & 0.050 \\
0.5, 0.5 & 21.60 & 14.40 & 0.992 & 0.956 & 0.005 & 0.048 \\
0.5, 0.8 & 21.43 & 14.40 & 0.990 & 0.964 & 0.005 & 0.052 \\
0.5, 1 & \textbf{21.33} & 14.40 & 0.989 & \textbf{0.968} & 0.005 & 0.047 \\
1, 0 & 22.88 & \textbf{14.35} & \textbf{0.994} & 0.933 & \textbf{0.004} & 0.052 \\
1, 0.3 & 22.02 & 14.37 & 0.993 & 0.951 & 0.005 & 0.052 \\
1, 0.5 & 21.84 & 14.39 & 0.992 & 0.957 & 0.005 & \textbf{0.046} \\
1, 0.8 & 21.61 & 14.43 & 0.991 & 0.964 & 0.005 & 0.052 \\
1, 1 & 21.59 & 14.45 & 0.990 & 0.967 & 0.005 & 0.057 \\
\bottomrule
\end{tabular}
\end{table}

%% file: tables/reference_settings.tex
\begin{table}[H]
\centering
\caption{Default training and sampling settings for the main EHRDiB completion and synthesis experiments.}
\label{tab:reference-settings}
\small
\begin{tabularx}{\linewidth}{@{}lX@{}}
\toprule
Component & Setting \\
\midrule
Latent and embedding dimensions & \(d_z=d_t=384\) \\
VAE widths & Encoder: 512-512-384; decoder: 384-512-512 \\
VAE dropout and value variance & 0.1; \(\sigma_{\mathrm{val}}^2=0.5\) \\
VAE optimization & AdamW; 50,000 updates; batch 2,048; learning rate \(5\times10^{-4}\); weight decay \(10^{-5}\) \\
KL regularization & Target \(10^{-5}\); linear warmup over 20,000 updates for MIMIC-IV, 10,000 for Duke \\
Endpoint network & Widths 1,024-512-384-384-384-512-1,024; dropout 0.05 \\
Bridge optimization & AdamW; 200,000 updates; batch 1,024; learning rate \(5\times10^{-4}\); weight decay 0; EMA 0.9999 \\
Checkpoint selection & Minimum validation objective; evaluate every 500 VAE / 1,000 bridge updates; EMA weights for bridge selection \\
Preconditioning & \(\nu_0=\nu_T=0.5\), \(\chi=0\), \(\kappa=250\) \\
Bridge schedule and training time & \(N=1{,}000\); symmetric schedule in Supplementary Methods~\ref{supp:training}; \(t\sim U(10^{-4},1)\) \\
Evidence-fraction bounds & Synthesis: \(U(0.1,0.5)\); completion: \(U(0.3,0.7)\) \\
Endpoint perturbation and stochasticity & Synthesis: 0.5 and 0.25; completion: 0 and 0.5 \\
Reverse grid & 128 positive grid points for MIMIC-IV; 64 for Duke; \(\rho=7\), followed by zero \\
\bottomrule
\end{tabularx}
\end{table}